\documentclass{article}
\usepackage[utf8]{inputenc}
\usepackage{textcomp}

\usepackage[main, final]{neurips_2026}

\usepackage{times}
\usepackage{amsmath}
\DeclareMathOperator{\argmax}{arg\,max}
\usepackage{mathtools}
\usepackage{amssymb}
\usepackage{hyperref}
\usepackage{url}
\usepackage{graphicx}
\usepackage{booktabs}
\usepackage{multirow}
\usepackage{pifont}
\usepackage{wrapfig}
\usepackage{xcolor}
\usepackage{enumitem}

\newcommand{\cmark}{\ding{51}}
\newcommand{\xmark}{\ding{55}}

\title{Differentiable Retrieval-Augmented Generation for Predicting Cellular Responses to Gene Perturbation}

\author{%
  Andrea Giuseppe Di Francesco\thanks{Equal contribution.} \\
  Sapienza University of Rome \\
  ISTI-CNR, Pisa, Italy \\
  \texttt{difrancesco@diag.uniroma1.it} \\
  \And
  Andrea Rubbi\footnotemark[1] \\
  University of Cambridge \\
  Wellcome Sanger Institute \\
  \texttt{ar2232@cam.ac.uk} \\
  \And
  Rishabh Jain \\
  University of Cambridge \\
  \texttt{rj412@cst.cam.ac.uk} \\
  \And
  Pietro Li\`{o} \\
  University of Cambridge \\
  \texttt{pl219@cam.ac.uk} \\
}

\begin{document}
\maketitle

\begin{abstract}
Predicting transcriptional responses to genetic perturbations is fundamental to functional genomics and therapeutic discovery. Recent deep learning models have shown promise in single-cell perturbation response prediction, but they typically generate each response in isolation, without explicitly leveraging related perturbations. We introduce \textbf{PT-RAG} (\textbf{P}erturbation-aware \textbf{T}wo-stage \textbf{R}etrieval-\textbf{A}ugmented \textbf{G}eneration), a plug-in retrieval-and-conditioning module for generative cellular perturbation response. PT-RAG augments an existing perturbation-response backbone with learned access to related perturbation contexts. The key challenge is that relevance is not fixed in this setting: functionally related genes may elicit different effects across cell types. PT-RAG addresses this with a two-stage retrieval mechanism: GenePT-based semantic retrieval first identifies $K$ candidate perturbations, after which a differentiable Gumbel-Softmax selector adaptively selects retrieved contexts conditioned on the control cell state, the query perturbation, and each candidate perturbation. We evaluate PT-RAG on two backbones, a STATE-style generator used as a frozen random reservoir and a fully trained scGPT, across cross-cell-type and cross-perturbation generalization tasks. PT-RAG consistently improves distributional similarity and often overall predictive quality; for example, on scGPT cross-cell-type results, the 2-Wasserstein distance drops by 5.9\% relative to scGPT alone. The code to reproduce our experiments is available at \url{https://github.com/difra100/PT-RAG_NIPS}.
\end{abstract}

\section{Introduction}
\label{sec:intro}

Understanding how cells respond to genetic perturbations is a fundamental challenge in systems biology with profound implications for drug discovery, disease modelling, and gene therapy~\citep{norman2019exploring}.
High-throughput Perturb-seq technologies now enable measurement of single-cell transcriptomic responses to thousands of genetic perturbations across diverse cell contexts~\citep{norman2019exploring, REPLOGLE20222559}, generating rich datasets that capture the complex landscape of cellular responses.
However, the combinatorial explosion of possible perturbations and cell contexts makes comprehensive experimental characterization infeasible, motivating the development of computational methods to predict perturbation responses \textit{in silico}.

Recent deep learning methods span a broad range of approaches: latent-space perturbation models~\citep{lotfollahi2019scgen, lotfollahi2021compositional}, optimal transport frameworks~\citep{bunne2023learning}, graph neural networks for gene-gene interactions~\citep{roohani2024predicting}, and transformer architectures that model cell populations as sequences~\citep{adduri2025state, klein2025cellflow}.
Despite their diversity, all of these methods share a critical architectural limitation: predictions are generated solely from the control cell state and perturbation identity, \textit{without leveraging knowledge about related perturbations} already observed in the training data.
This limitation is most acute in two practically important scenarios: predicting responses that are only partially observed during training, and predicting responses to perturbations that have never been experimentally tested.

In natural language processing, Retrieval-Augmented Generation (RAG)~\citep{lewis2021retrievalaugmentedgenerationknowledgeintensivenlp} has emerged as a powerful paradigm for incorporating external knowledge into generation.
In such settings, retrieval can rely on off-the-shelf text or image encoders~\citep{song2020mpnetmaskedpermutedpretraining, radford2021learningtransferablevisualmodels}, and even non-differentiable retrieval yields strong results because the notion of ``relevant context'' is well-defined and task-aligned.

\textbf{RAG beyond language models.}
Extending RAG to cellular biology is non-trivial for two reasons specific to this domain.
(1)~\textbf{No established retrieval metric}: there is no consensus on how to measure perturbation similarity.
One-hot encodings carry no semantic information; embedding-based similarities (e.g., via GenePT~\citep{genept}) capture functional descriptions but not cellular effects.
(2)~\textbf{Cell-type agnosticism}: standard retrieval depends only on the query perturbation, providing identical context regardless of whether predictions are made for T-cells, hepatocytes, or leukemia cells, though the same perturbation can have qualitatively different effects across these contexts.
These challenges raise a critical question: \textit{can RAG improve generation when the retrieval objective must itself be learned?}

\textbf{Our contributions.}
\begin{itemize}[leftmargin=*]
  \item We introduce \textbf{PT-RAG} (Perturbation-aware Two-stage Retrieval-Augmented Generation), the first differentiable RAG framework for cellular perturbation response prediction. PT-RAG uses a two-stage pipeline that first retrieves $K$ candidate perturbations via cosine similarity in GenePT embedding space, then applies Gumbel-Softmax~\citep{gumbel_softmax1, gumbel_softmax2} selection conditioned on both the target cell state and perturbation identity, enabling end-to-end optimization of what context benefits generation. We demonstrate that na\"{i}ve retrieval \emph{actively harms} performance, underscoring that differentiability and cell-type-aware conditioning are both necessary.
  \item We demonstrate that PT-RAG is a \textbf{plug-and-play} retrieval module rather than a backbone-specific architecture: in addition to the STATE family, we provide cross-cell-type transfer experiments on the scGPT perturbation generator~\citep{cui2023scGPT} (see Section~\ref{sec:scgpt_results}).
  \item We introduce a \textbf{cross-perturbation generalisation} protocol---a challenging split in which 60\% of perturbations are held out for testing, that tests the most practically relevant scenario: predicting responses to genes never experimentally perturbed. Together with cross-cell-type generalisation, the two protocols provide complementary evidence for when and why retrieval helps in cellular biology.
  \item We provide quantitative evidence that PT-RAG learns cell-type-specific retrieval patterns ($\sim$19\% perturbation overlap across cell types for the same query gene via Jaccard similarity analysis), and deliver statistically rigorous comparisons across both evaluation settings.
\end{itemize}

\section{Related Works}
\label{sec:related}

\paragraph{Perturbation response prediction.}
Existing perturbation prediction methods~\citep{lotfollahi2019scgen, lotfollahi2021compositional, bunne2023learning, roohani2024predicting, adduri2025state, klein2025cellflow} generate responses from the cell state and perturbation identity, but do not retrieve knowledge from related perturbations.
Namely, scGen~\citep{lotfollahi2019scgen} and CPA~\citep{lotfollahi2021compositional} learn perturbation effects in latent space, CellOT~\citep{bunne2023learning} frames prediction as neural optimal transport, STATE~\citep{adduri2025state} models cell populations as sequences with distributional losses, and CellFlow~\citep{klein2025cellflow} adds conditional flow matching.
scGPT~\citep{cui2023scGPT} provides a strong transformer backbone that we extend with RAG, while GEARS~\citep{roohani2024predicting} uses a GO co-annotation graph to propagate perturbation effects.
Our work extends the existing perturbation modelling paradigm from a generation-only perspective to a retrieval-augmented generation perspective, enabling contextual perturbation information to be used at prediction time.
Extended comparisons appear in Appendix~\ref{sec:extended_related}.

\paragraph{Differentiable RAG.}
Standard RAG~\citep{lewis2021retrievalaugmentedgenerationknowledgeintensivenlp} relies on fixed retrieval, while stochastic and differentiable variants~\citep{stochasticrag, gao-etal-2025-rag} learn retrieval decisions end to end.
Stochastic RAG~\citep{stochasticrag} and D-RAG~\citep{gao-etal-2025-rag} show the value of optimized retrieval, but they assume pretrained generators and well-defined similarity metrics.
PT-RAG instead learns relevance from scratch in a cell-type-dependent, non-textual generation setting.

\paragraph{RAG in biology.}
GeneRAG~\citep{generag}, scRAG~\citep{yu-etal-2025-scrag}, and agenticRAG for single-cell analysis \citep{agenticrag_singlecell}, all use retrieval to supply annotations or protein-centric context.
None of these methods targets the generation of cellular response \emph{distributions} conditioned on perturbation and cell state.
PT-RAG bridges this gap by applying retrieval directly to perturbation-response generation in a learned, task-conditioned, cell-state-aware manner.
Extended discussion appears in Appendix~\ref{sec:extended_related}.

\section{Methods}
\label{sec:method}

We consider the task of predicting cellular responses to single-gene perturbations.
Let $\mathcal{X} \subseteq \mathbb{R}^G$ be the space of gene expression profiles over $G$ genes.
Given a population of control cells $\{x_i^{ctrl}\}_{i=1}^N \subseteq \mathcal{X}$ and a perturbation $p^{pert} \in \mathcal{P}$ (where $\mathcal{P}$ is the set of perturbations), the goal is to predict the distribution of perturbed cells $\{\hat{x}_i^{pert}\}_{i=1}^N$.

\begin{figure}[t]
  \centering
  \includegraphics[width=\textwidth]{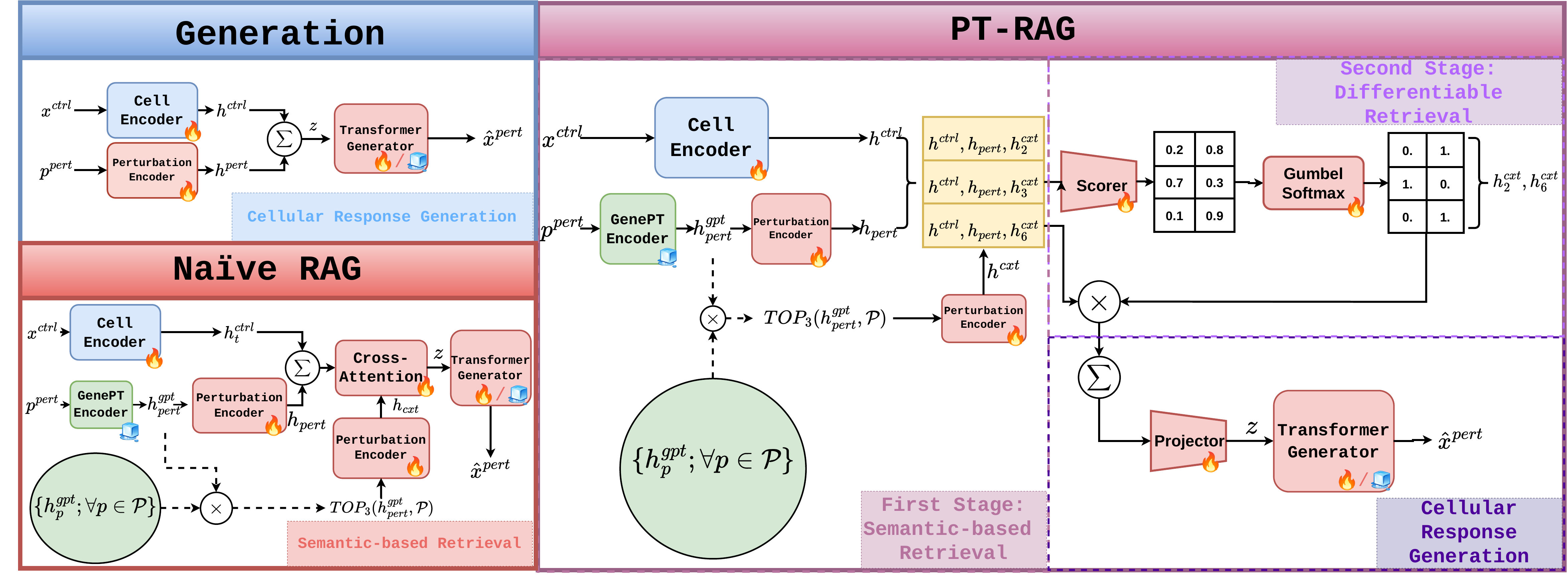}
  \caption{%
    \textbf{Comparison of architectures.}
    \textit{Left}: Generation baseline and Naïve RAG (non-differentiable retrieval, dotted lines; frozen components indicated by ice cube icon).
    \textit{Right}: PT-RAG uses two-stage differentiable retrieval with Gumbel-Softmax selection conditioned on $h^{ctrl}$, $h^{pert}$, and candidate embeddings $h_k^{cxt}$. The Transformer Generator is frozen in the STATE-based experiments and trained in the scGPT-based ones.
  }
  \label{fig:architecture}
\end{figure}

In this section we describe three approaches in increasing order of complexity: Generation baseline, Naïve RAG, and PT-RAG.
Figure~\ref{fig:architecture} provides a visual comparison of all architectures.

\subsection{Perturbation Representations with GenePT}
\label{sec:genept}

Unlike prior work that uses one-hot perturbation encodings~\citep{adduri2025state}, we leverage GenePT~\citep{genept} embeddings that capture semantic gene relationships.
For each gene~$g$, GenePT provides $h_g^{gpt} \in \mathbb{R}^{d_{gpt}}$ derived from GPT-3.5 encodings of the gene's NCBI description~\citep{ncbi} ($d_{gpt} = 1536$).
We construct a perturbation database $\mathcal{D} = \{h_p^{gpt} \mid p \in \mathcal{P}\}$ for all $|\mathcal{P}| \approx 2009$ perturbations and normalize embeddings to unit norm before similarity computation.

\subsection{Generation Baseline}
\label{sec:gen_baseline}
Following STATE~\citep{adduri2025state}, the Generation baseline uses:
\begin{equation}
  h^{ctrl} = \text{CellEncoder}(x^{ctrl}), \qquad h^{pert} = \text{PertEncoder}(h_{p}^{gpt}),
\end{equation}
combined as $z = h^{ctrl} + h^{pert}$ and decoded by a Transformer Generator:
$\hat{x}^{pert} = \text{TransformerGenerator}(z)$. 

The cell encoder is a trainable linear layer on the 2,000 HVG values. In the STATE experiments, the transformer generator and output projection are randomly initialised and frozen (reservoir~\citep{shen2021reservoir}); PertEncoder is a trainable single-layer MLP.
Predictions are made in isolation without leveraging related perturbation knowledge.
This decomposition is intentionally backbone-agnostic: to attach PT-RAG, a generator must expose (i) a control-state representation $h^{ctrl}$, (ii) a perturbation representation $h^{pert}$, and (iii) a differentiable point at which the retrieved context can condition the prediction of $\hat{x}^{pert}$. The retrieval logic itself is unchanged across backbones; only a small backbone-specific adapter is needed to connect it. In the STATE-based models, the context enters at the cell level, as the generator input, whereas in our scGPT instantiation~\citep{cui2023scGPT} it is added to the gene-token representations before the expression decoder (Appendix~\ref{sec:scgpt_impl}).

\subsection{Naïve RAG}

Given a perturbation query $h_{p}^{gpt}$, Naïve RAG retrieves its top-$K$ most similar perturbations via cosine similarity on GenePT embeddings:
$\mathcal{R}_{p^{pert}} = \mathrm{TOP}_K(h_{p}^{gpt}, \mathcal{D}) = \{p_{(1)}, \ldots, p_{(K)}\}$,
and integrates them via Cross-Attention:
\begin{equation}
  z = \text{CrossAttention}\!\left(q = h^{ctrl} + h^{pert},\; k = h^{cxt},\; v = h^{cxt}\right),
\end{equation}
where $h^{cxt} \in \mathbb{R}^{K \times d_h}$ is the matrix of the retrieved embeddings encoded with PertEncoder.
Retrieval is \textbf{non-differentiable} and depends \emph{only} on $h_{p}^{gpt}$, providing identical context regardless of cell type.

\subsection{PT-RAG: Perturbation-aware Two-stage Retrieval-Augmented Generation}
\label{sec:ptrag}

PT-RAG addresses the limitations of Naïve RAG through a two-stage differentiable retrieval pipeline.

\subsubsection{Stage 1: Semantic Candidate Retrieval}

Identical to Naïve RAG, Stage 1 non-differentiably selects $K$ candidates:
\begin{equation}
  \mathcal{R}_{p^{pert}} = \mathrm{TOP}_K(h_{p}^{gpt},\, \mathcal{D} \setminus \{h_{p}^{gpt}\})
    = \{p_{(1)}, \ldots, p_{(K)}\}.
\end{equation}
The query perturbation is excluded from the pool to prevent information leakage.
This stage efficiently prunes the search space from $|\mathcal{P}|$ to $K$ semantically relevant candidates.

\subsubsection{Stage 2: Differentiable Cell-Type-Aware Selection}

Each retrieved candidate $p_{(k)}$ is encoded as:
\begin{equation}
  h_k^{cxt} = \text{PertEncoder}(h_{p_{(k)}}^{gpt}) \in \mathbb{R}^{d_h}, \quad k = 1, \ldots, K.
\end{equation}
We form a \textbf{triplet} capturing the three-way relationship between cell state, target perturbation, and candidate:
\begin{equation}
  c_k = [h^{ctrl};\, h^{pert};\, h_k^{cxt}] \in \mathbb{R}^{3d_h},
\end{equation}
and score each candidate with a two-layer MLP:
\begin{equation}
  s_k = \mathrm{MLP}_{\mathrm{score}}(\mathrm{LayerNorm}(c_k)) \in \mathbb{R}^2,
\end{equation}
where $s_k$ outputs \texttt{exclude} and \texttt{include} logits.
We apply Straight-Through Gumbel-Softmax~\citep{gumbel_softmax1, gumbel_softmax2} to make $K$ \textbf{independent binary decisions} rather than a single selection among $K$ candidates:
\begin{equation}
  w_k = \mathrm{GumbelSoftmax}(s_k,\, \tau)[\texttt{include}] \in \{0, 1\}, \quad k = 1, \ldots, K.
\end{equation}
This allows multiple candidates to be selected simultaneously (or none).
Selected contexts are projected and aggregated:
\begin{equation}
  z = \sum_{k=1}^{K} w_k \cdot \mathrm{MLP}_{\mathrm{proj}}(c_k),
\end{equation}
and decoded: $\hat{x}^{pert} = \text{TransformerGenerator}(z)$.
\subsubsection{Training Objective}
We train all STATE-based models with the following objective (scGPT-based models use a mean-squared-error reconstruction loss in place of $\mathcal{L}_{\mathrm{dist}}$; Appendix~\ref{sec:scgpt_impl}).

\begin{equation}
  \mathcal{L} = \mathcal{L}_{\mathrm{dist}} + \lambda_{\mathrm{sparse}}\,\mathcal{L}_{\mathrm{sparse}},
  \label{eq:loss}
\end{equation}
where $\mathcal{L}_{\mathrm{dist}} = \mathrm{Energy}(\hat{x}^{pert}, x^{pert})$ is the energy distance~\citep{rizzo2016energy} and
$\mathcal{L}_{\mathrm{sparse}} = \frac{1}{K}\sum_{k=1}^{K} w_k$ penalizes dense masks, preventing mode collapse to always-retrieve-all.
The values of $\lambda_{\mathrm{sparse}}$ and $K$ used in each setting are reported in Appendix~\ref{sec:app_hyperparameters}; sensitivity analyses in Appendix~\ref{sec:sensitivity} confirm robustness to this choice.

Table~\ref{tab:method_comparison} summarizes the conceptual differences across all three methods.

\begin{table}[h]
  \caption{%
    \textbf{Method comparison.}
    All methods except PT-RAG use non-differentiable or no retrieval; PT-RAG uses two-stage differentiable selection.
  }
  \label{tab:method_comparison}
  \centering
  \small
  \setlength{\tabcolsep}{4pt}
  \begin{tabular}{lccc}
    \toprule
    & \textbf{Generation} & \textbf{Naïve RAG} & \textbf{PT-RAG} \\
    \midrule
    Retrieval              & \xmark & \cmark & \cmark \\
    Cell-type-aware Retrieval   & \xmark & \xmark & \cmark \\
    \bottomrule
  \end{tabular}
\end{table}

\section{Experimental Setup}
\label{sec:experiments}

\subsection{Dataset}

We use the Replogle-Nadig Perturb-seq dataset and evaluate two complementary generalization settings: cross-cell-type and cross-perturbation. In the cross-cell-type setting, we hold out each cell type in turn: about two-thirds of its perturbations are used as few-shot examples during training, and the rest are held out for validation and testing. In the cross-perturbation setting, all cell types are visible, and the perturbations themselves are split into train, validation, and test sets, with 60\% held out for testing and a dynamic retrieval pool that expands across stages.

\begin{figure}[t]
  \centering
  \begin{minipage}{0.42\textwidth}
    \centering
    \includegraphics[width=\linewidth]{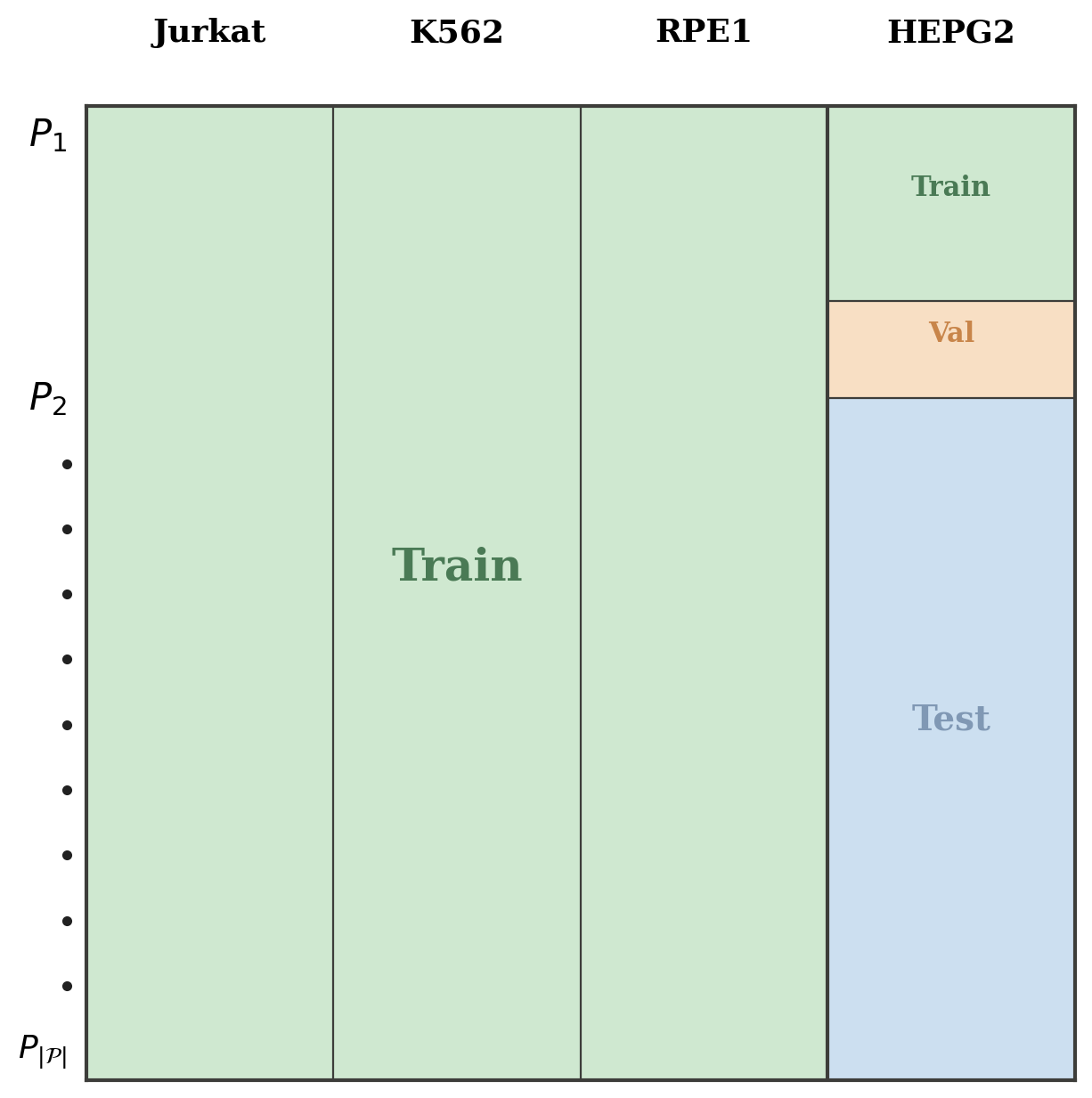}
    \caption{Cross-cell-type setting in Replogle-Nadig.}
    \label{fig:exp_settings1}
  \end{minipage}
  \hfill
  \begin{minipage}{0.42\textwidth}
    \centering
    \includegraphics[width=\linewidth]{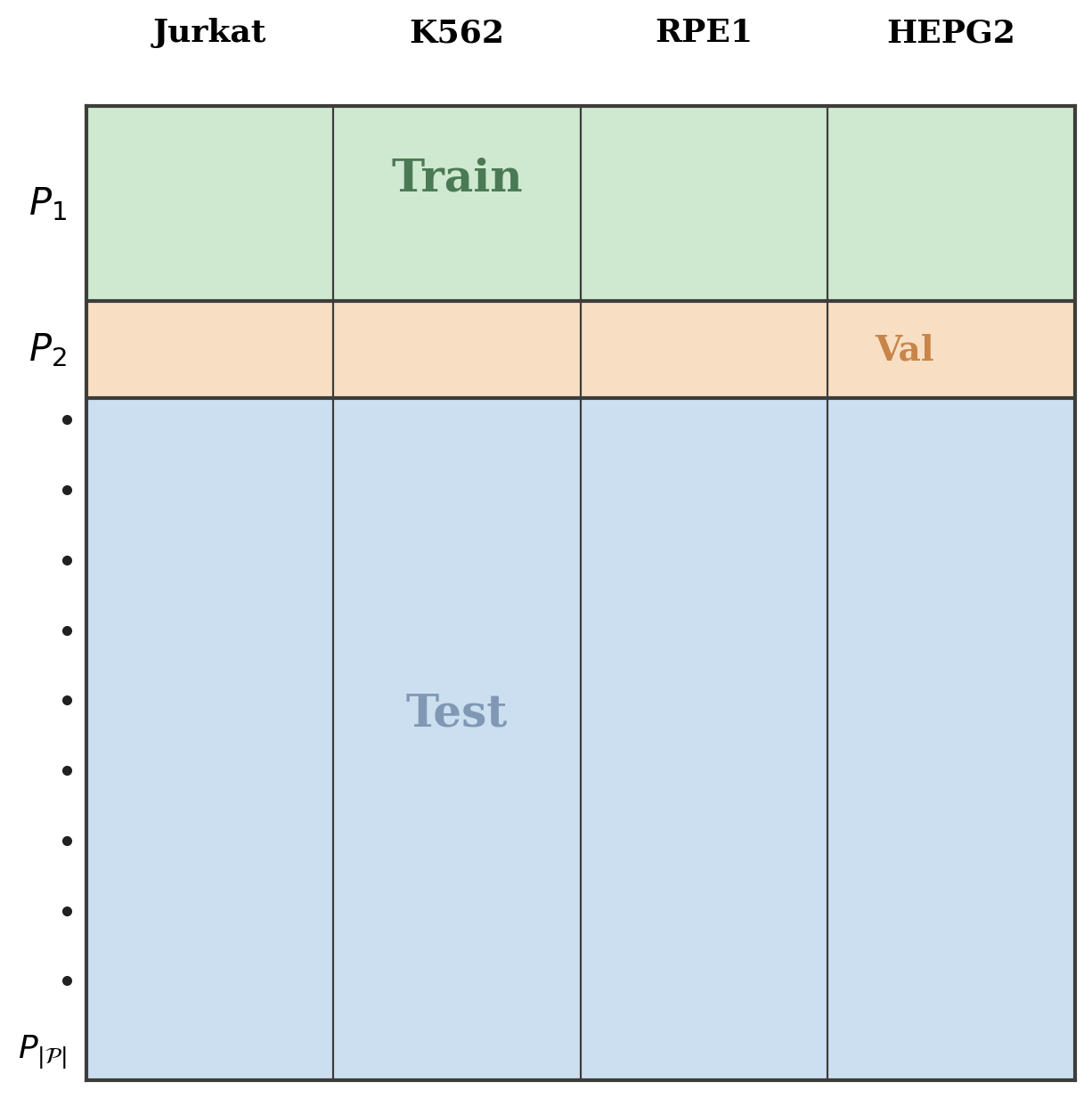}
    \caption{Cross-perturbation setting in Replogle-Nadig.}
    \label{fig:exp_settings2}
  \end{minipage}
\end{figure}

Figures~\ref{fig:exp_settings1} and~\ref{fig:exp_settings2} summarize the two settings. The full dataset description, split sizes, retrieval-pool protocol, and task details are provided in Appendix~\ref{sec:dataset_details}.

\subsection{Baselines and Models}

We compare four approaches for the main STATE-based experiments:
(1)~\textbf{STATE}~\citep{adduri2025state}: one-hot perturbation encoding, no retrieval;
(2)~\textbf{STATE+GenePT}: GenePT embeddings replace one-hot encodings, no retrieval;
(3)~\textbf{Naïve RAG}: top-$K = 32$ GenePT cosine retrieval, cross-attention integration, non-differentiable;
(4)~\textbf{PT-RAG} (ours): two-stage differentiable retrieval. In all STATE-based experiments, the generator is a frozen random reservoir (Section \ref{sec:gen_baseline}), whereas scGPT models are fully trained. Our STATE baselines therefore differ from the published STATE model and should not be read as a direct reproduction of it.
To evaluate backbone transferability, we additionally run a cross-cell-type study with an scGPT generator~\citep{cui2023scGPT} augmented by GenePT and PT-RAG (see Section \ref{sec:scgpt_results}).
Full architecture and training details are in Appendix~\ref{sec:impl_details}.
In the cross-perturbation setting, to avoid data leakage, the retrieval pool is stage-specific: $\mathcal{D}_{train}$ during training, $\mathcal{D}_{val}$ during validation, and all 2,009 perturbations at test time (excluding the query). We do not include the one-hot STATE baseline in this setting because one-hot perturbation codes are inherently transductive and cannot represent unseen perturbations in a cold-start regime. We also exclude Naïve RAG and compare PT-RAG against the strongest non-retrieval transferable baseline (STATE+GenePT).

\subsection{Evaluation Metrics}

According to common practices to assess gene expression generation quality \citep{rubbi2026standardizedframeworkevaluatinggene}, we use the following metrics (definitions in Appendix~\ref{sec:metrics}):
(1)~\textbf{Gene-level expression correlations} (Pearson, Spearman on differentially expressed genes [DEGs]);
(2)~\textbf{Expression reconstruction accuracy} (MSE$_{\mathrm{PCA50}}$);
(3)~\textbf{Distributional similarity in PCA space} ($W_1$, $W_2$, Energy distance).
Statistical significance is assessed via the Mann-Whitney U test with Benjamini-Hochberg FDR correction (details in Appendix~\ref{sec:statistical_details}).

\section{Results}
\label{sec:results}

\subsection{Cross-Cell-Type Generalization}
\label{sec:results_crosscell}

Table~\ref{tab:main_results} presents results across 1,635 test perturbations.

\begin{table}[t]
  \caption{%
    \textbf{Cross-cell-type generalization.}
    Mean values across 1,635 test perturbations from four cell types.
    Significance vs.\ PT-RAG: $\dagger$ ($p_{\mathrm{FDR}} < 0.01$), 
    $\dagger\dagger$ ($p_{\mathrm{FDR}} < 0.05$), 
    $\dagger\dagger\dagger$ ($p_{\mathrm{FDR}} < 0.1$).
    Bold: best.
    Results with standard deviations in Table~\ref{tab:main_results_full}.
  }
  \label{tab:main_results}
  \centering
  \small
  \setlength{\tabcolsep}{4pt}
  \begin{tabular}{lccc|c}
    \toprule
    \textbf{Metric} & \textbf{STATE} & \textbf{STATE+GenePT} & \textbf{Naïve RAG} & \textbf{PT-RAG} \\
    \midrule
    \multicolumn{5}{l}{\textit{Gene-level expression correlations}} \\
    Pearson DEG $\uparrow$ 
      & $0.624^{\dagger}$ & $0.631$ & $0.396^{\dagger}$ 
      & $\textbf{0.633}$ \\
    Spearman DEG $\uparrow$ 
      & $0.403^{\dagger}$ & $0.411$ & $0.307^{\dagger}$ 
      & $\mathbf{0.412}$ \\
    \midrule
    \multicolumn{5}{l}{\textit{Expression reconstruction accuracy}} \\
    MSE$_{\mathrm{PCA50}}$ $\downarrow$ 
      & $8.43$ & $8.42$ & $12.64^{\dagger}$ 
      & $\mathbf{8.39}$  \\
    \midrule
    \multicolumn{5}{l}{\textit{Distributional similarity (PCA space)}} \\
    $W_1$ $\downarrow$ 
      & $35.70^{\dagger}$ & $35.53^{\dagger\dagger\dagger}$ & $48.48^{\dagger}$ 
      & $\mathbf{35.41}$  \\
    $W_2$ $\downarrow$ 
      & $646.1^{\dagger}$ & $638.7^{\dagger\dagger}$ & $1189.5^{\dagger}$ 
      & $\mathbf{633.7}$ \\
    Energy $\downarrow$ 
      & $9.41^{\dagger\dagger\dagger}$ & $9.40$ & $14.18^{\dagger}$ 
      & $\mathbf{9.33}$  \\
    \bottomrule
  \end{tabular}
\end{table}

\textbf{Naïve RAG actively hurts performance.}
Despite incorporating retrieved context, Naïve RAG substantially underperforms STATE across all metrics (Pearson: 0.396 vs.\ 0.624; $W_2$: 1189.5 vs.\ 646.1).
This failure is a primary finding of this paper: in this domain, retrieval without differentiable, cell-type-aware selection is not merely unhelpful---it is actively harmful.
Ablations varying $K \in \{2, 5, 10, 32\}$ (Figures~\ref{fig:naïve_rag_k_ablation_core}--\ref{fig:naïve_rag_k_ablation_dist}) confirm that this failure is not due to insufficient retrieval breadth: even $K = 32$ Naïve RAG falls far short of STATE, with non-monotonic performance curves suggesting that larger pools introduce more noise rather than more signal.

\textbf{STATE+GenePT shows consistent but modest gains.}
Replacing one-hot encodings with GenePT embeddings yields small but consistent improvements over STATE (Pearson: 0.631 vs.\ 0.624), demonstrating the value of semantic gene representations.
The remaining gap to PT-RAG is concentrated in distributional metrics ($W_1$, $W_2$), suggesting that cell-type-specific context selection adds population-level fidelity beyond what semantic embeddings alone provide.

\textbf{PT-RAG.}
PT-RAG demonstrates statistically significant improvements over STATE in gene-level correlations (Pearson: $p_{\mathrm{FDR}} = 2.4 \times 10^{-8}$; Spearman: $p_{\mathrm{FDR}} = 4.9 \times 10^{-10}$), and distributional similarity ($W_1$: $p_{\mathrm{FDR}} = 3.4 \times 10^{-6}$; $W_2$: $p_{\mathrm{FDR}} = 5.5 \times 10^{-8}$).
Improvements over STATE+GenePT are more modest and concentrated in Wasserstein distances ($W_2$: $p_{\mathrm{FDR}} = 0.041$), consistent with the interpretation that retrieval adds cell-type-specific context selection on top of semantic encoding rather than semantic encoding itself.
Notably, PT-RAG does not consistently outperform STATE+GenePT on K562 (Table~\ref{tab:k562_results}), suggesting that retrieval benefits may be cell-type-dependent.

Overall, these results indicate that retrieved GenePT embeddings capture substantial functional relationships, and that end-to-end differentiable, cell-state-aware retrieval (PT-RAG) improves both gene-level correlations and distributional alignment relative to non-retrieval and naive retrieval baselines.

\subsubsection{Transfer to scGPT backbone}
\label{sec:scgpt_results}
The scGPT transfer section tests whether PT-RAG behaves as a backbone-agnostic retrieval module rather than a backbone-specific trick. The retrieval logic is unchanged: Stage 1 still retrieves GenePT candidates by cosine similarity and Stage 2 still makes cell-state-aware include/exclude decisions with Gumbel-Softmax; only the generator backbone changes.
For readability in the main text, we report means only; full mean $\pm$ standard deviation values are provided in Appendix Table~\ref{tab:scgpt_crosscell_full}.

\begin{table}[t]
  \caption{\textbf{Cross-cell-type generalization: Results on scGPT.}
    Mean values across 1,635 test perturbations from four cell types.
    Significance vs.\ PT-RAG: $\dagger$ ($p_{\mathrm{FDR}} < 0.01$), 
    $\dagger\dagger$ ($p_{\mathrm{FDR}} < 0.05$), 
    $\dagger\dagger\dagger$ ($p_{\mathrm{FDR}} < 0.1$).
    Bold: best.
    Results with standard deviations in Table~\ref{tab:scgpt_crosscell_full}.
  }
  \label{tab:scgpt_crosscell}
  \centering
  \small
  \setlength{\tabcolsep}{4pt}
  \renewcommand{\arraystretch}{1.1}
  \begin{tabular}{lcc | c}
    \toprule
    \textbf{Metric} 
    & \textbf{scGPT} 
    & \textbf{scGPT+GenePT} 
    & \textbf{scGPT+PT-RAG}  \\
    \midrule
     \multicolumn{4}{l}{\textit{Gene-level expression correlations}} \\
    Pearson DEG $\uparrow$ 
      & \textbf{0.9209} 
      & 0.9167
      & 0.9128 \\

    Spearman DEG $\uparrow$ 
      & 0.9500 $^\dagger$
      & \textbf{0.9509} $^\dagger$
      & 0.9368  \\
    \midrule
    \multicolumn{4}{l}{\textit{Expression reconstruction accuracy}} \\
    MSE$_{PCA50}$ $\downarrow$ 
      & 18.6440 $^\dagger$
      & 19.8907 $^\dagger$
      & \textbf{17.3212} \\
    \midrule
    \multicolumn{4}{l}{\textit{Distributional similarity (PCA space)}} \\
    
    $W_1$  $\downarrow$ 
      & 33.7992 $^\dagger$ 
      & 34.6746 $^\dagger$ 
      & \textbf{32.7828} \\
      
    $W_2$  $\downarrow$ 
      & 578.6169 $^\dagger$ 
      & 609.7348 $^\dagger$ 
      & \textbf{544.6977} \\
      
    Energy $\downarrow$ 
      & 23.5341 $^\dagger$ 
      & 24.3923 $^\dagger$ 
      & \textbf{22.3410}  \\

    \bottomrule
  \end{tabular}
\end{table}
\textbf{scGPT transfer: retrieval improves distributional fidelity while preserving backbone transferability.} The scGPT results mirror the main STATE findings on the distributional side: PT-RAG is best on all three geometry metrics ($W_1$, $W_2$, Energy) and on reconstruction (MSE$_{\mathrm{PCA50}}$), with statistically significant differences versus both baselines (Table~\ref{tab:scgpt_crosscell}). Relative to scGPT alone, PT-RAG reduces MSE$_{\mathrm{PCA50}}$ by 7.1\%, $W_1$ by 3.0\%, $W_2$ by 5.9\% (from 578.6 to 544.7) and Energy by 5.1\%; relative to scGPT+GenePT, the $W_2$ reduction is 10.7\%. These gains come with slightly lower gene-level correlations (Pearson $-0.008$, Spearman $-0.013$ versus scGPT; the Spearman difference is significant).

	\textbf{GenePT alone can degrade scGPT, while PT-RAG recovers and surpasses in most metrics.}
Compared with vanilla scGPT (one-hot perturbation encoding), scGPT+GenePT is worse on Pearson, MSE$_{\mathrm{PCA50}}$, $W_1$, $W_2$, and Energy (with only a small Spearman gain). PT-RAG then recovers from this drop and outperforms both scGPT and scGPT+GenePT on reconstruction and all distributional metrics, at the cost of slightly lower gene-level correlations. This pattern supports the same conclusion as in STATE: semantic embeddings are useful, but they need cell-state-aware differentiable selection to consistently improve generation quality.

\subsection{Cross-Perturbation Generalization}
\label{sec:results_crosspert}

Table~\ref{tab:crosspert_results} reports results averaged across three folds.
This setting is harder than cross-cell-type because 60\% of all perturbations are unseen during training, so the model must recover from a much smaller direct supervision signal and rely more heavily on the retrieval pool.
The table therefore serves as a stress test of whether PT-RAG still improves the geometry of the predicted response distribution when the perturbation split is much more severe.
For readability in the main text, we report fold means only; full mean $\pm$ standard deviation values are provided in Appendix Table~\ref{tab:crosspert_results_full}.

\begin{table}[t]
  \caption{%
    \textbf{Cross-perturbation generalization.}
    Results averaged across three folds. Significance vs.\ PT-RAG: $\dagger$ ($p_{\mathrm{FDR}} < 0.01$), 
    $\dagger\dagger$ ($p_{\mathrm{FDR}} < 0.05$), 
    $\dagger\dagger\dagger$ ($p_{\mathrm{FDR}} < 0.1$).
    Bold: best.
    Results with standard deviations in Table~\ref{tab:crosspert_results_full}.
  }
  \label{tab:crosspert_results}
  \centering
  \small
  \setlength{\tabcolsep}{4pt}
  \begin{tabular}{lcc}
    \toprule
    \textbf{Metric} & \textbf{STATE+GenePT} & \textbf{PT-RAG} \\
    \midrule
    \multicolumn{3}{l}{\textit{Gene-level expression correlations}} \\
    Pearson DEG $\uparrow$ & $\mathbf{0.6377}$ & $0.6314$ \\
    Spearman DEG $\uparrow$ & $\mathbf{0.4293}$ & $0.4224$ \\
    \midrule
    \multicolumn{3}{l}{\textit{Expression reconstruction accuracy}} \\
    MSE$_{\mathrm{PCA50}}$ $\downarrow$ & $1.1477^{\dagger}$ & $\mathbf{1.1316}$ \\
    \midrule
    \multicolumn{3}{l}{\textit{Distributional similarity (PCA space)}} \\
    $W_1$ $\downarrow$ & $0.8043^{\dagger}$ & $\mathbf{0.8016}$ \\
    $W_2$ $\downarrow$ & $0.6436^{\dagger}$ & $\mathbf{0.6367}$ \\
    Energy $\downarrow$ & $0.3385^{\dagger}$ & $\mathbf{0.3357}$ \\
    \bottomrule
  \end{tabular}
\end{table}

	\textbf{Cross-perturbation: PT-RAG consistently improves most metrics, with significant gains on reconstruction and distributional geometry.}
This setting is more challenging than cross-cell-type because 60\% of perturbations are never seen during training. In this regime, PT-RAG improves 4/6 reported metrics versus STATE+GenePT, and these gains are statistically significant on MSE$_{\mathrm{PCA50}}$, $W_1$, $W_2$, and Energy (all marked $\dagger$ in Table~\ref{tab:crosspert_results}). The strongest pattern is again distributional: PT-RAG yields lower Wasserstein and Energy distances, indicating better recovery of population-level response structure for unseen genes.

	\textbf{Correlation behavior is complementary, not contradictory.}
STATE+GenePT remains slightly higher on Pearson and Spearman DEG, while PT-RAG is better on reconstruction and distributional metrics. This is consistent with retrieval acting primarily as a geometric/population-level correction mechanism rather than a pure rank-correlation optimizer.

	\textbf{Why one-hot baselines are excluded here.}
A one-hot perturbation representation cannot encode perturbations absent from training: for unseen genes, the model has no transferable perturbation embedding and therefore no meaningful cold-start initialization. This is exactly why cross-perturbation evaluation is centered on transferable representations (GenePT) and why PT-RAG is evaluated against STATE+GenePT in this setting.

The dynamic retrieval pool is a distinctive feature of this setting.
At test time, the model has access to all the 2,009 characterized perturbations (excluding the query), whereas during training only the train split is retrievable.
A control in which the test-time pool is restricted to the training perturbations gives essentially the same results (Appendix~\ref{sec:additional_controls}), so the gains are not explained by the larger pool available at test time.

\subsection{Generalization to a Different Perturbation Modality}
\label{sec:norman}

The experiments above use a single dataset family based on CRISPRi knockdown. To test whether PT-RAG transfers to a different perturbation modality, we evaluate it on the Perturb-seq dataset of~\citet{norman2019exploring}, in which genes are \emph{activated} with CRISPRa in \texttt{K562} cells. Since the dataset contains a single cell type, only the cross-perturbation setting applies: the 105 single-gene perturbations are split into 32 train, 10 validation and 63 test perturbations, with the same stage-dependent retrieval pool as in Section~\ref{sec:crossperturbation}. We use the STATE-based reservoir setting and report one fold; details are in Appendix~\ref{sec:norman_setup}.

\begin{table}[h]
\centering
\caption{\textbf{Cross-perturbation generalization on Norman et al.\ (CRISPRa).} Mean values across the 63 test perturbations of one fold. Significance vs.\ PT-RAG: $\dagger$ ($p_{\mathrm{FDR}} < 0.01$), $\dagger\dagger$ ($p_{\mathrm{FDR}} < 0.05$). \textbf{Bold}: best.}
\label{tab:norman}
\small
\begin{tabular}{lcc}
\toprule
\textbf{Metric} & \textbf{STATE+GenePT} & \textbf{PT-RAG} \\
\midrule
\multicolumn{3}{l}{\textit{Gene-level expression correlations}} \\
Pearson DEG $\uparrow$  & 0.2985$^{\dagger}$ & \textbf{0.3856} \\
Spearman DEG $\uparrow$ & 0.2760$^{\dagger}$ & \textbf{0.3564} \\
\midrule
\multicolumn{3}{l}{\textit{Expression reconstruction accuracy}} \\
MSE$_{\mathrm{PCA50}}$ $\downarrow$ & 5.4482 & \textbf{5.2871} \\
\midrule
\multicolumn{3}{l}{\textit{Distributional similarity (PCA space)}} \\
$W_1$ $\downarrow$   & 1.3706$^{\dagger}$ & \textbf{1.3339} \\
$W_2$ $\downarrow$   & 3.0658$^{\dagger\dagger}$ & \textbf{2.9649} \\
Energy $\downarrow$  & 0.7985$^{\dagger\dagger}$ & \textbf{0.7815} \\
\bottomrule
\end{tabular}
\end{table}

\textbf{PT-RAG improves all metrics on Norman.} Table~\ref{tab:norman} shows that PT-RAG outperforms STATE+GenePT on every reported metric, and does so in each of the 63 test perturbations. In contrast to Replogle-Nadig, where the gains are concentrated in distributional metrics, the largest improvements here are in gene-level correlations (Pearson $+0.087$, Spearman $+0.080$), while MSE$_{\mathrm{PCA50}}$, $W_1$, $W_2$ and Energy decrease by 2.1--3.3\%. The differences are statistically significant on all metrics except MSE$_{\mathrm{PCA50}}$. This suggests that the lower DEG correlations observed in the Replogle-Nadig cross-perturbation setting are specific to that setting rather than intrinsic to retrieval. We note that this experiment uses a small perturbation set and a single fold, so it should be read as supporting evidence rather than as a full replication.

\subsection{Cell-Type-Specific Retrieval Patterns}
\label{sec:cell_type_retrieval_patterns}

To validate that PT-RAG's differentiable retrieval genuinely conditions on cellular context, we analyze the overlap in selected perturbations across cell types using Jaccard similarity.

\begin{wrapfigure}{r}{0.47\textwidth}
  \vspace{-8pt}
  \centering
  \includegraphics[width=0.45\textwidth]{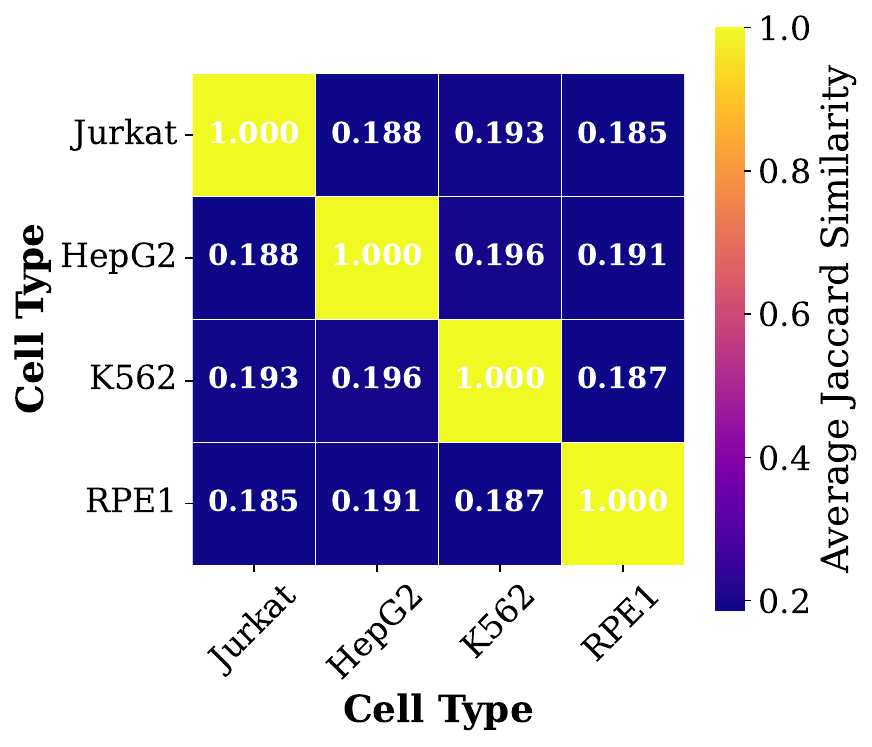}
  \caption{Jaccard similarity of top-10 retrieved perturbations across cell types. Off-diagonal values (range 0.185--0.196, mean 0.191) confirm that only $\sim$19\% of selected perturbations overlap between cell types for the same query gene.}
  \label{fig:jaccard}
  \vspace{-8pt}
\end{wrapfigure}

For each of the 33 genes common to all four cell types in the cross-cell-type test set, we extract the top-10 most frequently selected perturbations by the Gumbel-Softmax mechanism in each cell type and compute pairwise Jaccard similarity.
Figure~\ref{fig:jaccard} shows the resulting matrix.

\textbf{Low overlap confirms cell-type-specific retrieval.}
Off-diagonal Jaccard values average 0.191, indicating that $\sim$19\% of selected perturbations overlap between cell types for the same query gene.
This low overlap confirms that PT-RAG actively conditions on cell state rather than relying solely on the query perturbation---a capability that Naïve RAG fundamentally lacks.

\textbf{Biologically coherent selection.}
Despite variation across cell types, PT-RAG maintains functional coherence.
For WARS (tryptophanyl-tRNA synthetase), it retrieves other aminoacyl-tRNA synthetases across all cell types, but with different specific members (Jurkat: EARS2, DARS, VARS; HepG2: SARS2, GART, TARS; K562: FARSB, KARS, FARS2; RPE1: KARS, GART, TARS, QARS).
This is consistent with known cell-type-specific differences in protein synthesis demands, though we note these are suggestive observations rather than rigorous biological claims.
Additional retrieval examples appear in Appendix~\ref{sec:retrieval_examples}.

\section{Discussion}
\label{sec:discussion}

\textbf{Cross-cell-type results.}
Table~\ref{tab:main_results} gives the cleanest picture in the paper: Naïve RAG is consistently worse than the no-retrieval baseline, STATE+GenePT gives a small improvement over STATE, and PT-RAG is best on every metric reported there.
These results suggest that retrieval helps when it is learned end to end and when the selector can choose contexts that match the current cell state.

\textbf{scGPT transfer.}
The scGPT transfer table shows two complementary facts. First, PT-RAG is not tied to one generator: the same two-stage retrieval policy transfers to scGPT without changing retrieval formulation. Second, the gains are concentrated where retrieval should matter most: reconstruction and distributional geometry. In particular, $W_2$ drops from 578.6 (scGPT) to 544.7 (scGPT+PT-RAG), with consistent improvements in $W_1$ and Energy as well.
The table also highlights that replacing one-hot perturbation encoding with GenePT alone can reduce performance in this backbone; PT-RAG recovers this drop and outperforms both baselines on most metrics.

\textbf{Cross-perturbation results.}
The cross-perturbation experiments provide complementary evidence: it tests retrieval in the strictest cold-start regime, where unseen perturbations dominate the test split.
In this setting, the STATE one-hot baseline is not applicable, so comparison against STATE+GenePT isolates the added value of retrieval over the strongest transferable non-retrieval baseline.
PT-RAG improves most reported metrics and is significant on reconstruction and all distributional rows, which is exactly the regime where context selection should help: preserving the geometry of unseen perturbed populations. The Norman experiment (Section~\ref{sec:norman}) shows the same benefit under a different perturbation modality, with gains also in gene-level correlations.

\textbf{Why the method works.}
The tables point to a consistent interpretation.
GenePT already gives the model a useful semantic prior, but Naïve RAG shows that retrieval by itself is not enough; the retrieved contexts have to be filtered by a differentiable, cell-aware selector.
PT-RAG improves the parts of the prediction problem that depend on population structure, which is why the retrieval module is best understood as a control mechanism over context rather than as a replacement for the generator.

\textbf{Limitations.}
PT-RAG adds computation relative to the baselines because retrieval and selection are learned on top of the generator, so the method is not free. Our STATE-based experiments use a frozen, randomly initialised generator; we do not claim that the reported gains carry over to a fully trained generator, which we leave for future work. In the cross-perturbation setting, the retrieval pool at test time contains the embeddings of all perturbations, including those never seen in training; we regard this as part of the protocol rather than as an advantage of the method. Moreover, Stage~1 retrieval is cell-type agnostic, so a relevant perturbation that is not among the top-$K$ GenePT neighbours cannot be recovered by Stage~2 (see the controls in Appendix~\ref{sec:additional_controls}).
The current study is also limited to single-gene perturbations, which keeps the biological setting focused but leaves combinatorial perturbations and other modalities that would require a completely new analysis on context modelling for future work.
Finally, the retrieval examples are biologically suggestive rather than causal, so they should be read as model behavior rather than mechanistic proof.

\section{Conclusion}
\label{sec:conclusion}

We have presented PT-RAG as a retrieval module for perturbation biology that is both differentiable and reusable across backbones.
The cross-cell-type results show the main claim most cleanly: learned retrieval beats Naïve RAG and improves on the retrieval-free baseline, with the strongest gains appearing where population geometry matters.

The scGPT experiment reinforces the plug-and-play framing, because the same retrieval rule can be attached to another generator without changing the retrieval logic. The cross-perturbation results then show how the same method behaves under a harder split in which direct perturbation supervision is much thinner.

That setting matters because it isolates whether retrieval can still help when the model has to lean more heavily on related contexts rather than on direct examples of the query perturbation.

Taken together, we conclude that PT-RAG is effective when retrieval is learned, cell-aware, and transferred as a module rather than baked into one backbone.

\begin{ack}
We thank the anonymous reviewers and the area chair for their constructive feedback and insightful discussions, which helped us improve this work. We thank Muhammad Shoaib Malik for an independent reproduction effort that helped us clarify the experimental setting.

\textbf{Funding.} A.G.D.F. was partially supported by the European Union's Horizon Europe Research and Innovation programme under the RenAI project (GA No.~101291378), and by the ENRESFOOD project, funded under the FutureFoodS partnership. The project is co-funded by the Austrian Science Fund (FWF; 10.55776/KIN4661525), the Dutch Ministry of Agriculture, Fisheries, Food Security and Nature (BO-43-219-026), the Italian Ministry of University and Research (MUR), the German Federal Ministry of Research, Technology and Space (031B1717), and the European Union's Horizon Europe research and innovation programme via FutureFoodS, the European Partnership for a Sustainable Future of Food Systems (GA No.~101136361).

\textbf{Competing interests.} The authors declare no competing interests.
\end{ack}

\bibliographystyle{unsrtnat}
\bibliography{neurips_2026_CR}

\newpage
\appendix
\section*{Appendix Overview}

\begin{itemize}[leftmargin=*]
  \item \textbf{Appendix \ref{sec:gumbel_softmax}}: Gumbel-Softmax mathematical background.
  \item \textbf{Appendix \ref{sec:metrics}}: Evaluation metric definitions.
  \item \textbf{Appendix \ref{sec:impl_details}}: Full implementation and hyperparameter details.
  \item \textbf{Appendix \ref{sec:dataset_details}}: Description about datasets, and task, namely cross-cell-type and cross-perturbation, and the additional Norman dataset.
  \item \textbf{Appendix \ref{sec:extended_related}}: Extended related work.
  \item \textbf{Appendix \ref{sec:extended_results}}: Extended results: cell-type-specific retrieval examples, full tables with standard deviations, per-cell-type breakdowns, sensitivity analyses, additional controls, and computational costs.
  \item \textbf{Appendix \ref{sec:statistical_details}}: Statistical test details.
  \item \textbf{Appendix \ref{sec:llm_usage}}: LLM usage statement.
\end{itemize}

\section{Additional Preliminaries: Gumbel-Softmax}
\label{sec:gumbel_softmax}

\subsection{The Gumbel Distribution}

\begin{figure}[h]
  \centering
  \includegraphics[width=0.45\textwidth]{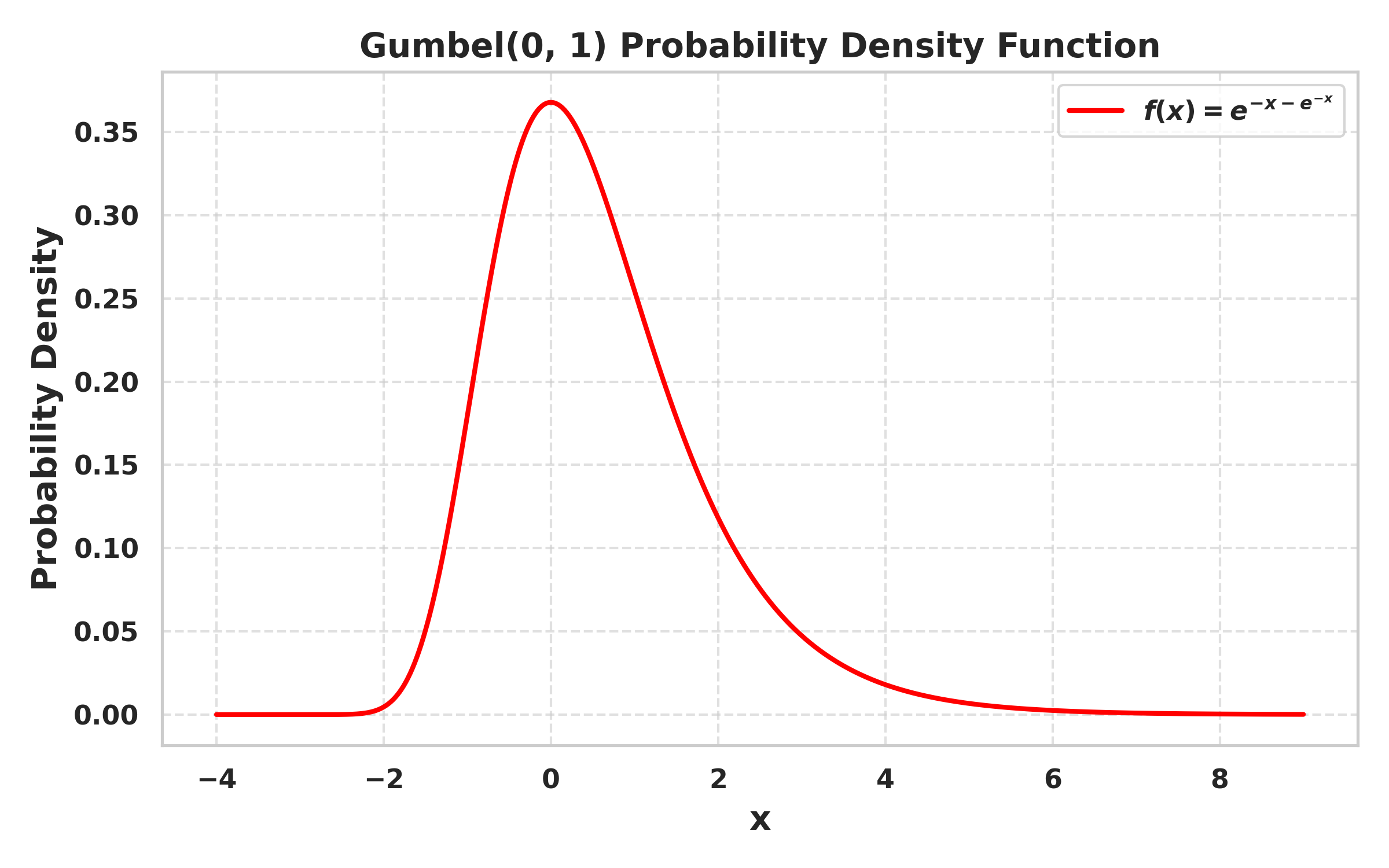}
  \caption{PDF of Gumbel$(0,1)$: $f(x) = e^{-x - e^{-x}}$.}
  \label{fig:gumbel_pdf}
\end{figure}

The Gumbel$(0,1)$ distribution has PDF:
\begin{equation}
  f(x) = e^{-x - e^{-x}},
\end{equation}
illustrated in Figure~\ref{fig:gumbel_pdf}.

\subsection{The Gumbel-Softmax Trick}

Given logits $\pi = (\pi_1, \ldots, \pi_n)$, the Gumbel-Softmax~\citep{gumbel_softmax1, gumbel_softmax2} provides a continuous, differentiable approximation to categorical sampling:
\begin{equation}
  y_i = \frac{\exp((\log \pi_i + g_i)/\tau)}{\sum_{j} \exp((\log \pi_j + g_j)/\tau)},
  \quad g_i \overset{\mathrm{iid}}{\sim} \mathrm{Gumbel}(0,1),
\end{equation}
where $\tau > 0$ is temperature. As $\tau \to 0$, the distribution approaches a one-hot (categorical) sample.

\subsection{Straight-Through Estimator}

For hard binary decisions, we use the Straight-Through Gumbel-Softmax estimator.
Forward pass applies $\argmax$ to obtain a hard one-hot; backward pass computes gradients through the soft probabilities:
\begin{equation}
  \mathbf{y} = \mathbf{y}^{\mathrm{hard}} + \bigl(\mathbf{y}^{\mathrm{soft}} - \mathrm{sg}(\mathbf{y}^{\mathrm{soft}})\bigr),
\end{equation}
where $\mathbf{y}^{\mathrm{hard}} = \mathrm{one\_hot}(\argmax_i y_i^{\mathrm{soft}})$ and $\mathrm{sg}(\cdot)$ is stop-gradient.
This enables end-to-end training of models with discrete selection.

\textbf{PT-RAG formulation.}
The Gumbel-Softmax is applied independently to each of the $K$ candidates (binary include/exclude), not as a softmax over $K$ candidates.
This allows multiple candidates to be selected simultaneously or none at all, giving the scoring MLP full flexibility.

\section{Evaluation Metrics}
\label{sec:metrics}

\subsection{Gene-Level Expression Correlations}

For each test condition, DEGs are identified by comparing the observed perturbed
cells with the control cells: a gene is called differentially expressed if its
Welch's $t$-test $p$-value is at most $0.1$, and its absolute $\log_2$ fold change
is at least $0.05$. Correlations are then computed on these genes:
\begin{itemize}
  \item \textbf{Pearson correlation}: linear relationship between predicted and observed expression changes across DEGs.
  \item \textbf{Spearman correlation}: rank-order correlation of expression changes; robust to outliers.
\end{itemize}

\subsection{Expression Reconstruction Accuracy}

\begin{itemize}
  \item $\mathrm{MSE}_{\mathrm{PCA50}}$: mean squared error in a 50-dimensional PCA
space, fit on the observed expression profiles used for evaluation.

\end{itemize}

\subsection{Distributional Similarity}

Both predicted and true populations are projected to 50-dimensional PCA space before computing:
\begin{itemize}
  \item \textbf{Wasserstein distances $W_1$, $W_2$}~\citep{rubner2000emd, cuturi2013sinkhorn}: optimal transport distances, sensitive to both shape and support.
  \item \textbf{Energy distance}~\citep{rizzo2016energy}: $\mathcal{E}(P, Q) = 2\mathbb{E}[\|X\!-\!Y\|] - \mathbb{E}[\|X\!-\!X'\|] - \mathbb{E}[\|Y\!-\!Y'\|]$; primary training objective.
\end{itemize}

\section{Implementation Details}
\label{sec:impl_details}

\subsection{Shared Architecture}

All STATE-based models use a Llama-style~\citep{touvron2023llamaopenefficientfoundation} transformer generator (8 layers, hidden size 128, 12 attention heads, feed-forward size 2{,}784) operating on cell populations of sequence length 64, with batch size 64.

\textbf{Reservoir setting.}
In all STATE-based experiments the transformer generator and its linear output projection are \emph{randomly initialised and kept frozen}; no pretrained weights are used. The generator therefore acts as a fixed random reservoir, in the spirit of reservoir computing and reservoir transformers~\citep{shen2021reservoir}. Only the components around it are trained: the cell encoder, the perturbation encoder, the retrieval modules (where present), and the readout that maps the generator output to gene expression.

\textbf{Cell Encoder.}
A trainable linear layer maps expression profiles over $G = 2000$ highly variable genes to 128-dimensional representations.

\textbf{Perturbation Encoder.}
Trainable single-layer MLP: $\mathbb{R}^{1536} \to \mathbb{R}^{128}$.

\textbf{Readout.}
A trainable MLP (hidden sizes 1024, 1024, 512) maps the generator output to the $G$ expression values used for evaluation.

\textbf{Parameter counts.}
Of the 20.97M parameters of PT-RAG, 5.21M are trainable and 15.76M are frozen (11.41M transformer layers, 0.26M output projection, and a 4.10M token-embedding table that is unused because inputs are provided as embeddings). The corresponding trainable counts are 5.18M (STATE), 5.11M (STATE+GenePT) and 5.36M (Na\"ive RAG); the frozen part is identical across methods.

\subsection{PT-RAG-Specific Components}
\label{sec:app_hyperparameters}
\begin{itemize}
  \item \textbf{Scoring MLP} ($\mathrm{MLP}_{\mathrm{score}}$): two-layer network, hidden dim 128; input $[h^{ctrl}; h^{pert}; h_k^{cxt}] \in \mathbb{R}^{384}$; output $\mathbb{R}^2$.
  \item \textbf{Projection MLP} ($\mathrm{MLP}_{\mathrm{proj}}$): one-layer network, hidden dim 128.
  \item Gumbel temperature $\tau = 0.5$. Cross-cell-type (STATE): $K = 32$, $\lambda_{\mathrm{sparse}} = 0.1$. Cross-perturbation (STATE): $K = 128$, $\lambda_{\mathrm{sparse}} = 1.0$; the controls in Appendix~\ref{sec:additional_controls} use a model with $K = 32$. Norman: $K = 16$, $\lambda_{\mathrm{sparse}} = 1.0$.

\end{itemize}

\subsection{scGPT Backbone}
\label{sec:scgpt_impl}

To test transfer beyond STATE, we implement PT-RAG on top of an scGPT perturbation generator~\citep{cui2023scGPT} (12 transformer layers, embedding size 512, 8 heads). Each cell is represented as a sequence of up to 512 gene tokens; every token combines a gene embedding, an encoding of the continuous control expression value, and a perturbation-flag embedding. A decoder predicts the perturbed expression of each gene from the transformer output.

\textbf{Initialisation and training.}
Gene embeddings, the value encoder and the transformer are initialised from the public scGPT checkpoint where parameter names match; the attention projection weights do not match and are trained from scratch. \emph{All} parameters are trained (no component is frozen), with learning rate $5\times10^{-5}$, batch size 64 and 25{,}000 steps. scGPT models are trained with a mean-squared-error reconstruction loss, to which PT-RAG adds the sparsity term of the training objective.

\textbf{scGPT+GenePT.}
The GenePT embedding of the query perturbation is projected to the model dimension, $h^{pert} = \mathrm{MLP}(h^{gpt}_p) \in \mathbb{R}^{512}$, and added to the transformer output at every gene position before decoding.

\textbf{scGPT+PT-RAG.}
The retrieval module is the same as in the STATE instantiation. Stage~1 retrieves the top-$K$ candidates by GenePT cosine similarity. For Stage~2, the control-cell representation $h^{ctrl}$ is read from the transformer output (first sequence position), each candidate is encoded with the same GenePT projection, and the triplets $[h^{ctrl}; h^{pert}; h^{cxt}_k]$ are scored with a two-layer MLP followed by Straight-Through Gumbel-Softmax, as in the STATE instantiation. The selected projected triplets are summed and the resulting context vector is added to the transformer output at every gene position, on top of $h^{pert}$, before the expression decoder. Only the interface differs from STATE: the context conditions gene-token representations after the transformer, rather than replacing the generator input. We use $K=128$ and $\lambda_{\mathrm{sparse}} = 1.0$.

\subsection{Training Configuration}

\begin{itemize}
    \item STATE-based models: Adam optimizer, learning rate $10^{-3}$; up to 50{,}000 steps, validation every 2{,}000 steps.
  \item scGPT-based models: learning rate $5 \times 10^{-5}$, 25{,}000 steps (Appendix~\ref{sec:scgpt_impl}).
  \item Hardware: NVIDIA A100 (80 GB). PT-RAG convergence: $\approx$8--10 hours per target cell type.
\end{itemize}

\subsection{Cross-Perturbation Protocol Details}

Folds are constructed by random partition of the 2,009 perturbations into 670 train / 133 validation / 1,206 test perturbations (approximately 33\% / 7\% / 60\%).
Three independent folds are generated; final results are averaged across folds.

GenePT embeddings of the perturbations that become available in the pool at validation and test time are used for retrieval indexing (Stage 1).

\section{Dataset and Tasks}
\label{sec:dataset_details}

\subsection{Dataset}

We evaluate on the Replogle-Nadig dataset~\citep{REPLOGLE20222559, Nadig2024.07.03.601903}, a large-scale Perturb-seq study with 643{,}413 single-cell transcriptomic profiles.
The data span four human cell lines: \texttt{K562} (chronic myeloid leukemia)~\citep{k562}, \texttt{Jurkat} (T-cell lymphocyte)~\citep{jurkat}, \texttt{RPE1} (retinal pigment epithelial-1)~\citep{rpe1}, and \texttt{HepG2} (hepatocellular carcinoma)~\citep{hepg2}.
The dataset contains 2{,}023 unique single-gene perturbations and a non-targeting control; we restrict retrieval and evaluation to the 2{,}009 perturbations with an available GenePT embedding.
Not every perturbation is profiled in every cell line: the number of perturbations per cell line ranges from 1{,}340 (\texttt{HepG2}) to 1{,}537 (\texttt{Jurkat}).
Each cell is represented by $G = 2{,}000$ highly variable genes. An additional dataset used to test a different perturbation modality is described in Section \ref{sec:norman_setup}.

\subsection{Evaluation Setting 1: Cross-Cell-Type Generalization}
\label{sec:crosscelltype_setup}

Each cell type serves as a held-out target in turn, using the split files released with STATE~\citep{adduri2025state}.
Models are trained on all perturbation responses from the three remaining cell types.
For the target cell type, about two thirds of its perturbations (900--1{,}014, depending on the cell type) are provided as few-shot examples during training; the remaining perturbations are held out, with 60--78 used for validation and 380--449 (about 28\%) for testing.
Perturbations held out in the target cell type remain available in the source cell types in which they were profiled.
After restricting to perturbations with a GenePT embedding, the test sets contain 375 (\texttt{HepG2}), 401 (\texttt{K562}), 443 (\texttt{Jurkat}) and 416 (\texttt{RPE1}) perturbations, for a total of $n = 1{,}635$.

This setting tests generalization across \emph{biological contexts}: given perturbation data from related cell types, can the model predict responses in a partially observed cell type?
It is the primary setting for evaluating cell-type-awareness in PT-RAG.

\subsection{Evaluation Setting 2: Cross-Perturbation Generalization}
\label{sec:crossperturbation}

This setting tests generalization across the \emph{perturbation space} itself: predicting cellular responses to perturbations never seen during training.

\textbf{Protocol.}
All four cell types are always visible during training, validation, and testing, so no cell type is held out.
The 2{,}009 perturbations are randomly split into \textbf{670 train}, \textbf{133 validation}, and \textbf{1{,}206 test} perturbations (approximately 33\% / 7\% / 60\%).
This ratio is challenging by design, simulating realistic experimental campaigns in which most genes have not yet been perturbed.
Because a perturbation is evaluated separately in each cell line where it was profiled, each fold yields about 3{,}430 test cell-type--perturbation pairs.

The retrieval pool is dynamic and stage-dependent (the query perturbation is always excluded):
\begin{itemize}[leftmargin=*]
  \item \textbf{Training}: only train-split perturbations are available for retrieval ($|\mathcal{D}_{\mathrm{train}}| = 669$).
  \item \textbf{Validation}: train and validation perturbations are available ($|\mathcal{D}_{\mathrm{val}}| = 802$).
  \item \textbf{Testing}: all perturbations are available for retrieval, including test-split perturbations ($|\mathcal{D}_{\mathrm{test}}| = 2{,}008$).
\end{itemize}
The pool contains the GenePT embeddings of these perturbations, not their measured expression responses.
This dynamic pool reflects the scenario where retrieval can draw on the description of any gene of interest at inference time.
Three independent folds are generated and results averaged.

The two task setups are visualized in Figures~\ref{fig:exp_settings1}, \ref{fig:exp_settings2} in the main text; here we keep the full protocol details.

\textbf{Relationship to cross-cell-type setting.}
The two settings are complementary.
Cross-cell-type asks: \textit{can retrieval help when the perturbation is known but the cell type is only partially observed?}
Cross-perturbation asks: \textit{can retrieval help when the cell type is known but the perturbation is unseen?}
The cross-perturbation setting is the more direct test of the value of retrieval: when a query perturbation has no direct training signal, the representations of functionally related perturbations seen during training may provide the main source of perturbation-specific information.
It also represents a practically relevant experimental scenario: in discovery campaigns, one typically has transcriptomic data across cell types but wants to predict the effect of genes that have never been perturbed.


\subsection{Additional Dataset: \citet{norman2019exploring}}
\label{sec:norman_setup}

To test PT-RAG beyond the Replogle-Nadig data, we use the Perturb-seq dataset of~\citet{norman2019exploring}, which differs in perturbation modality: genes are \emph{activated} with CRISPRa, whereas Replogle-Nadig uses CRISPRi knockdown.
The dataset contains a single cell line (\texttt{K562}).
We keep the single-gene perturbations and the control cells, giving 69{,}570 cells and 105 single-gene perturbations.
Expression counts are normalized to $10^4$ counts per cell and log-transformed, and genes are aligned to the gene set of the Replogle-Nadig data (6{,}350 of its 6{,}642 genes are present in the Norman data; missing genes are set to zero).

Since there is only one cell type, only the cross-perturbation setting applies.
We use the same protocol as in Section~\ref{sec:crossperturbation}: the 105 perturbations are randomly split into 32 train, 10 validation and 63 test perturbations (approximately 30\% / 10\% / 60\%).
One perturbation has no GenePT embedding and is excluded from retrieval, so the retrieval pool contains 31 perturbations during training, 41 during validation and 104 at test time (the query is always excluded).
Because the training pool is small, we use $K = 16$ candidates and $\lambda_{\mathrm{sparse}} = 1.0$, and train for 10{,}000 steps; all other settings are as in the Replogle-Nadig experiments.
Results are reported for one fold.

\section{Extended Related Work}
\label{sec:extended_related}

\subsection{Single-Cell Perturbation Response Modeling}

Computational prediction of single-cell perturbation responses has advanced rapidly in recent years.
scGen~\citep{lotfollahi2019scgen} learns additive perturbation vectors in a VAE latent space, while CPA~\citep{lotfollahi2021compositional} introduces compositional modeling with covariate disentanglement.
CellOT~\citep{bunne2023learning} casts perturbation response prediction as neural optimal transport between control and perturbed distributions, and STATE~\citep{adduri2025state} models cell populations as sequences with distributional losses such as energy distance and Sinkhorn divergence.
CellFlow~\citep{klein2025cellflow} extends this family with conditional flow matching for trajectory prediction.
scGPT~\citep{cui2023scGPT} provides a strong transformer backbone for single-cell modeling and is a natural candidate to extend with RAG for perturbation prediction, which is exactly the transfer setting studied in the main text.

A complementary line of work leverages gene-gene interaction graphs.
GEARS~\citep{roohani2024predicting} constructs gene regulatory graphs from GO co-annotation with Jaccard-weighted edges and uses graph neural networks via neural message passing~\citep{gilmer2017neuralmessagepassingquantum} to propagate perturbation effects, enabling combinatorial knockout prediction.

Across these methods, the shared limitation is that predictions are generated from cell state and perturbation identity alone, without explicitly retrieving knowledge from related perturbations.
Our framework extends STATE~\citep{adduri2025state} and scGPT~\citep{cui2023scGPT} with retrieval-augmented generation so contextual perturbation information can be used at prediction time rather than compressed into a direct perturbation representation.
This makes PT-RAG a retrieval module for perturbation biology rather than a replacement for the underlying generator.

\subsection{Retrieval-Augmented Generation}

RAG~\citep{lewis2021retrievalaugmentedgenerationknowledgeintensivenlp} combines parametric knowledge stored in model weights with non-parametric knowledge retrieved from external sources.
In its standard form, retrieval is fixed and query-based: the generator is conditioned on retrieved content, but the retrieval step itself is not learned end to end.
Stochastic RAG~\citep{stochasticrag} and D-RAG~\citep{gao-etal-2025-rag} show that optimized retrieval can improve performance, but they assume pretrained generators and well-defined similarity metrics.
PT-RAG instead learns relevance from scratch in a cell-type-dependent, non-textual generation setting where both the retrieved content and the output are distributions over cells rather than text.

RAG has also expanded beyond text to images, video, audio, and graphs, showing that the retrieval principle is not inherently language-bound.
The key difference in our setting is that the retrieved items are perturbation embeddings and the generator output is a predicted cellular population, which requires task-conditioned retrieval rather than lexical or document matching.

\subsection{RAG in Biology}

Within biology, most RAG systems retrieve literature, annotations, or structured context for downstream tasks.
GeneRAG~\citep{generag} and scRAG~\citep{yu-etal-2025-scrag} retrieve textual context for biological question answering, and related systems such as agenticRAG for single-cell analysis and RetrievalAugmentedProteinEncoder focus on annotation-centric or protein-centric retrieval.
These approaches demonstrate that retrieval is useful in biology, but none addresses generation of cellular response distributions conditioned on perturbation and cell state.

The closest conceptual overlap with PT-RAG is retrieval-augmented encoding, where retrieved context is injected into a learned representation before prediction.
That perspective helps explain why our method transfers cleanly to scGPT: the same retrieval policy can be attached to a transformer backbone as a plug-in module.
PT-RAG is therefore the first RAG framework for cellular response modeling that jointly handles learned retrieval, cell-state conditioning, and non-textual generation.

Overall, the appendix literature supports the same conclusion as the main paper: existing perturbation-response models and existing biological RAG systems each solve part of the problem, but PT-RAG is the first to combine them in a differentiable, cell-aware retrieval-and-generation framework.

\subsection{Differentiable Retrieval in Other Domains}

Stochastic RAG~\citep{stochasticrag} formulates retrieval as sampling without replacement using Gumbel-Top-K, demonstrating benefits of end-to-end optimization on diverse NLP benchmarks.
D-RAG~\citep{gao-etal-2025-rag} applies differentiable retrieval to knowledge graph QA, jointly optimizing a subgraph retriever and LLM-based generator via Gumbel-Softmax reparameterization.
Both works operate in domains with pretrained generators and a relatively clear notion of relevance.
PT-RAG adapts the differentiable retrieval idea to a domain where these assumptions fail: no pretrained retriever exists, relevant context is cell-type-dependent, and the generator produces cell distributions rather than text.

\section{Extended Results}
\label{sec:extended_results}

\subsection{Cell-Type-Specific Retrieval Examples}
\label{sec:retrieval_examples}

We present top-5 most frequently selected perturbations across cell types for five illustrative query genes.

\textbf{WARS (Tryptophanyl-tRNA Synthetase).}
Jurkat: EARS2, DARS, SEPSECS, CTU2, VARS;
HepG2: SARS2, GART, KARS, EARS2, TARS;
K562: FARSB, KARS, FARS2, EPRS, DARS;
RPE1: KARS, GART, TARS, QARS, AARS2.
All are aminoacyl-tRNA synthetases, reflecting functional coherence with the query gene's role in translation; specific members vary by cell type, consistent with differential protein synthesis demands across T-cells, hepatocytes, myeloid, and epithelial contexts.

\textbf{DDX27 (DEAD-Box RNA Helicase).}
Jurkat: DHX15, GEMIN4, DHX16, DDX19A, DDX23;
HepG2: EDC4, DDX56, DHX36, GEMIN4, DDX1;
K562: DDX11, DDX1, DDX6, DHX36, DHX33;
RPE1: EDC4, DDX10, DDX56, DHX29, DDX19A.
Selection of distinct DDX/DHX family members suggests cell-type-specific RNA processing pathway utilization; GEMIN4 is shared between Jurkat and HepG2, potentially reflecting shared splicing demands.

\textbf{MRPL39 (Mitochondrial Ribosomal Protein L39).}
Jurkat: MRPL23, MRPL50, MRPL51, MRPL37, BRIP1;
HepG2: RPL13, MRPL20, MRPL17, MRPL4, MRPL33;
K562: MRPL36, MRPL20, MRPL51, MRPL54, MRPL4;
RPE1: MRPL10, MRPL3, MRPL37, MRPL51, MRPL35.
HepG2's inclusion of cytoplasmic RPL13 may reflect the high metabolic demands of hepatocytes coordinating mitochondrial and cytoplasmic translation.

\textbf{RPS2 (Ribosomal Protein S2).}
Jurkat: RPS17, RPS28, RPL35, RPS3A, RPS15;
HepG2: RPL35, RPS10, RPS16, RPS17, RPL4;
K562: RPS24, RPL30, RPS17, RPS9, RPL4;
RPE1: RPS3A, RPS5, RPS28, RPL9, RPS10.
The mixture of large (RPL) and small (RPS) subunit proteins varies by cell type, potentially reflecting cell-type-specific ribosome stoichiometry.

\textbf{TPRKB (TP53RK Binding Protein).}
Jurkat: PGAM5, ING3, PTK2, PKMYT1, DBF4;
HepG2: TTK, PLK1, ING3, PGAM5, GAK;
K562: DBF4, PLK1, BRK1, PTK2, PRKRA;
RPE1: CDK2, GTF3C4, GAK, BRK1, PLK1.
Cell-cycle kinase preferences differ across cell types (PKMYT1 in Jurkat, CDK2 in RPE1), suggestive of context-dependent cell cycle regulatory requirements.

\textbf{Summary.}
These examples demonstrate that PT-RAG's differentiable retrieval learns biologically plausible, cell-type-specific patterns: functional coherence (retrieving genes from the same family or pathway) is maintained while specific context selection adapts to the cellular background.
We note these observations are suggestive and not a rigorous biological validation.

\subsection{Cross-Cell-Type Results with Standard Deviations}

Table~\ref{tab:main_results_full} presents complete cross-cell-type results.

\begin{table}[t]
  \caption{\textbf{Cross-cell-type: full results with standard deviations.}
    $n = 1{,}635$ (375 HepG2, 416 RPE1, 443 Jurkat, 401 K562).}
  \label{tab:main_results_full}
  \centering\small
  \begin{tabular}{lcccc}
    \toprule
    \textbf{Metric} & \textbf{STATE} & \textbf{STATE+GenePT} & \textbf{Naïve RAG} & \textbf{PT-RAG} \\
    \midrule
    Pearson DEG $\uparrow$ & $0.624 \pm 0.048$ & $0.631 \pm 0.051$ & $0.396 \pm 0.063$ & $\mathbf{0.633 \pm 0.048}$ \\
    Spearman DEG $\uparrow$ & $0.403 \pm 0.046$ & $0.411 \pm 0.052$ & $0.307 \pm 0.041$ & $\mathbf{0.412 \pm 0.051}$ \\
    \midrule
    MSE$_{\mathrm{PCA50}}$ $\downarrow$ & $8.43 \pm 0.93$ & $8.42 \pm 0.88$ & $12.64 \pm 1.04$ & $\mathbf{8.39 \pm 0.87}$ \\
    \midrule
    $W_1$ $\downarrow$ & $35.70 \pm 1.76$ & $35.53 \pm 1.68$ & $48.48 \pm 1.71$ & $\mathbf{35.41 \pm 1.62}$ \\
    $W_2$ $\downarrow$ & $646.1 \pm 63.6$ & $638.7 \pm 60.6$ & $1189.5 \pm 83.3$ & $\mathbf{633.7 \pm 58.4}$ \\
    Energy $\downarrow$ & $9.41 \pm 1.18$ & $9.40 \pm 1.15$ & $14.18 \pm 1.17$ & $\mathbf{9.33 \pm 1.15}$ \\
    \bottomrule
  \end{tabular}
\end{table}

\subsection{Cross-Cell-Type Results on scGPT with Standard Deviations}

Table~\ref{tab:scgpt_crosscell_full} provides the full cross-cell-type scGPT results (mean $\pm$ standard deviation), corresponding to Table~\ref{tab:scgpt_crosscell} in the main text.

\begin{table}[t]
  \caption{\textbf{Cross-cell-type generalization on scGPT: full results with standard deviations.}
    Mean values across 1,635 test perturbations from four cell types.
    Significance vs.\ PT-RAG: $\dagger$ ($p_{\mathrm{FDR}} < 0.01$), 
    $\dagger\dagger$ ($p_{\mathrm{FDR}} < 0.05$), 
    $\dagger\dagger\dagger$ ($p_{\mathrm{FDR}} < 0.1$).
    Bold: best.
  }
  \label{tab:scgpt_crosscell_full}
  \centering
  \small
  \setlength{\tabcolsep}{4pt}
  \renewcommand{\arraystretch}{1.1}
  \begin{tabular}{lcc | c}
    	\toprule
    	\textbf{Metric} 
    & \textbf{scGPT} 
    & \textbf{scGPT+GenePT} 
    & \textbf{scGPT+PT-RAG}  \\
    \midrule
    \multicolumn{4}{l}{\textit{Gene-level expression correlations}} \\
    Pearson DEG $\uparrow$ 
      & \textbf{0.9209 $\pm$ 0.0158} 
      & 0.9167 $\pm$ 0.0149
      & 0.9128 $\pm$ 0.0146 \\

    Spearman DEG $\uparrow$ 
      & 0.9500 $\pm$ 0.0341
      & \textbf{0.9509 $\pm$ 0.0306}
      & 0.9368 $\pm$ 0.0299  \\
    \midrule
    \multicolumn{4}{l}{\textit{Expression reconstruction accuracy}} \\
    MSE$_{PCA50}$ $\downarrow$ 
      & 18.6440 $\pm$ 2.0080 $^\dagger$
      & 19.8907 $\pm$ 2.8476 $^\dagger$
      & \textbf{17.3212 $\pm$ 2.1105} \\
    \midrule
    \multicolumn{4}{l}{\textit{Distributional similarity (PCA space)}} \\
    $W_1$  $\downarrow$ 
      & 33.7992 $\pm$ 1.6254 $^\dagger$ 
      & 34.6746 $\pm$ 2.2780 $^\dagger$ 
      & \textbf{32.7828 $\pm$ 1.8324} \\
      
    $W_2$  $\downarrow$ 
      & 578.6169 $\pm$ 54.8356 $^\dagger$ 
      & 609.7348 $\pm$ 80.0043 $^\dagger$ 
      & \textbf{544.6977 $\pm$ 60.1570} \\
      
    Energy $\downarrow$ 
      & 23.5341 $\pm$ 1.7914 $^\dagger$ 
      & 24.3923 $\pm$ 2.0678 $^\dagger$ 
      & \textbf{22.3410 $\pm$ 1.7828}  \\

    \bottomrule
  \end{tabular}
\end{table}

\subsection{Cross-Perturbation Results with Standard Deviations}

Table~\ref{tab:crosspert_results_full} provides the full cross-perturbation results (mean $\pm$ standard deviation), corresponding to Table~\ref{tab:crosspert_results} in the main text.

\begin{table}[t]
  \caption{%
    	\textbf{Cross-perturbation generalization: full results with standard deviations.}
    Results averaged across three folds. Significance vs.\ PT-RAG: $\dagger$ ($p_{\mathrm{FDR}} < 0.01$), 
    $\dagger\dagger$ ($p_{\mathrm{FDR}} < 0.05$), 
    $\dagger\dagger\dagger$ ($p_{\mathrm{FDR}} < 0.1$).
    Bold: best.
  }
  \label{tab:crosspert_results_full}
  \centering
  \small
  \setlength{\tabcolsep}{4pt}
  \begin{tabular}{lcc}
    	\toprule
    	\textbf{Metric} & \textbf{STATE+GenePT} & \textbf{PT-RAG} \\
    \midrule
    \multicolumn{3}{l}{\textit{Gene-level expression correlations}} \\
    Pearson DEG $\uparrow$ & $\mathbf{0.6377 \pm 0.0367}$ & $0.6314 \pm 0.0394$ \\
    Spearman DEG $\uparrow$ & $\mathbf{0.4293 \pm 0.0344}$ & $0.4224 \pm 0.0381$ \\
    \midrule
    \multicolumn{3}{l}{\textit{Expression reconstruction accuracy}} \\
    MSE$_{\mathrm{PCA50}}$ $\downarrow$ & $1.1477 \pm 0.2243^{\dagger}$ & $\mathbf{1.1316 \pm 0.2150}$ \\
    \midrule
    \multicolumn{3}{l}{\textit{Distributional similarity (PCA space)}} \\
    $W_1$ $\downarrow$ & $0.8043 \pm 0.0582^{\dagger}$ & $\mathbf{0.8016 \pm 0.0578}$ \\
    $W_2$ $\downarrow$ & $0.6436 \pm 0.1134^{\dagger}$ & $\mathbf{0.6367 \pm 0.1093}$ \\
    Energy $\downarrow$ & $0.3385 \pm 0.0459^{\dagger}$ & $\mathbf{0.3357 \pm 0.0442}$ \\
    \bottomrule
  \end{tabular}
\end{table}

\subsection{Per-Cell-Type Results}

In this section, we want to provide the reader with information at the cell-level granularity, namely, how the model performs cell-perturbation pairs for each different training. We have results where the held-out cell was any among \texttt{HepG2}, \texttt{RPE1}, \texttt{Jurkat}, and \texttt{K562}. Such results are in Tables \ref{tab:hepg2_results}, \ref{tab:rpe1_results}, \ref{tab:jurkat_results}, \ref{tab:k562_results}. We see how the patterns of Table \ref{tab:main_results_full}, reflects the results from the specific cell-lines.

\begin{table}[t]
  \caption{Results on \texttt{HepG2} ($n = 375$).}
  \label{tab:hepg2_results}
  \centering\small
  \begin{tabular}{lcccc}
    \toprule
    \textbf{Metric} & \textbf{STATE} & \textbf{STATE+GenePT} & \textbf{Naïve RAG} & \textbf{PT-RAG} \\
    \midrule
    Pearson DEG $\uparrow$ & $0.596 \pm 0.051$ & $0.595 \pm 0.055$ & $0.351 \pm 0.054$ & $\mathbf{0.604 \pm 0.049}$ \\
    Spearman DEG $\uparrow$ & $0.398 \pm 0.044$ & $0.394 \pm 0.058$ & $0.289 \pm 0.036$ & $\mathbf{0.401 \pm 0.049}$ \\
    MSE$_{\mathrm{PCA50}}$ $\downarrow$ & $8.89 \pm 0.32$ & $8.86 \pm 0.32$ & $13.35 \pm 0.77$ & $\mathbf{8.83 \pm 0.31}$ \\
    $W_1$ $\downarrow$ & $36.30 \pm 1.34$ & $35.53 \pm 1.20$ & $48.68 \pm 1.31$ & $\mathbf{35.41 \pm 1.18}$ \\
    $W_2$ $\downarrow$ & $668.5 \pm 49.5$ & $637.7 \pm 43.3$ & $1198.5 \pm 64.5$ & $\mathbf{631.8 \pm 42.3}$ \\
    Energy $\downarrow$ & $9.68 \pm 0.69$ & $9.54 \pm 0.67$ & $14.25 \pm 0.90$ & $\mathbf{9.48 \pm 0.67}$ \\
    \bottomrule
  \end{tabular}
\end{table}

\begin{table}[t]
  \caption{Results on \texttt{RPE1} ($n = 416$).}
  \label{tab:rpe1_results}
  \centering\small
  \begin{tabular}{lcccc}
    \toprule
    \textbf{Metric} & \textbf{STATE} & \textbf{STATE+GenePT} & \textbf{Naïve RAG} & \textbf{PT-RAG} \\
    \midrule
    Pearson DEG $\uparrow$ & $0.621 \pm 0.043$ & $0.633 \pm 0.043$ & $0.450 \pm 0.052$ & $\mathbf{0.640 \pm 0.042}$ \\
    Spearman DEG $\uparrow$ & $0.419 \pm 0.035$ & $0.432 \pm 0.033$ & $0.342 \pm 0.027$ & $\mathbf{0.438 \pm 0.033}$ \\
    MSE$_{\mathrm{PCA50}}$ $\downarrow$ & $9.56 \pm 0.35$ & $9.48 \pm 0.32$ & $13.29 \pm 0.80$ & $\mathbf{9.44 \pm 0.33}$ \\
    $W_1$ $\downarrow$ & $36.75 \pm 1.25$ & $36.62 \pm 1.24$ & $49.31 \pm 1.46$ & $\mathbf{36.53 \pm 1.22}$ \\
    $W_2$ $\downarrow$ & $682.5 \pm 47.1$ & $676.3 \pm 46.4$ & $1227.5 \pm 72.9$ & $\mathbf{672.8 \pm 45.8}$ \\
    Energy $\downarrow$ & $10.17 \pm 0.84$ & $10.14 \pm 0.84$ & $14.56 \pm 0.81$ & $\mathbf{10.08 \pm 0.84}$ \\
    \bottomrule
  \end{tabular}
\end{table}

\begin{table}[t]
  \caption{Results on \texttt{Jurkat} ($n = 443$).}
  \label{tab:jurkat_results}
  \centering\small
  \begin{tabular}{lcccc}
    \toprule
    \textbf{Metric} & \textbf{STATE} & \textbf{STATE+GenePT} & \textbf{Naïve RAG} & \textbf{PT-RAG} \\
    \midrule
    Pearson DEG $\uparrow$ & $0.636 \pm 0.048$ & $\mathbf{0.652 \pm 0.049}$ & $0.398 \pm 0.056$ & $0.649 \pm 0.048$ \\
    Spearman DEG $\uparrow$ & $0.392 \pm 0.056$ & $0.424 \pm 0.058$ & $0.318 \pm 0.033$ & $\mathbf{0.424 \pm 0.054}$ \\
    MSE$_{\mathrm{PCA50}}$ $\downarrow$ & $\mathbf{7.35 \pm 0.41}$ & $7.36 \pm 0.41$ & $11.46 \pm 0.62$ & $7.36 \pm 0.41$ \\
    $W_1$ $\downarrow$ & $35.94 \pm 1.52$ & $36.10 \pm 1.50$ & $49.06 \pm 1.50$ & $\mathbf{35.82 \pm 1.48}$ \\
    $W_2$ $\downarrow$ & $654.4 \pm 56.7$ & $660.1 \pm 56.1$ & $1219.6 \pm 75.4$ & $\mathbf{649.9 \pm 55.1}$ \\
    Energy $\downarrow$ & $8.15 \pm 1.01$ & $8.23 \pm 1.02$ & $13.12 \pm 1.10$ & $\mathbf{8.14 \pm 1.01}$ \\
    \bottomrule
  \end{tabular}
\end{table}

\begin{table}[t]
  \caption{Results on \texttt{K562} ($n = 401$).}
  \label{tab:k562_results}
  \centering\small
  \begin{tabular}{lcccc}
    \toprule
    \textbf{Metric} & \textbf{STATE} & \textbf{STATE+GenePT} & \textbf{Naïve RAG} & \textbf{PT-RAG} \\
    \midrule
    Pearson DEG $\uparrow$ & $\mathbf{0.639 \pm 0.037}$ & $0.638 \pm 0.037$ & $0.381 \pm 0.046$ & $0.633 \pm 0.041$ \\
    Spearman DEG $\uparrow$ & $\mathbf{0.406 \pm 0.040}$ & $0.391 \pm 0.044$ & $0.274 \pm 0.029$ & $0.383 \pm 0.047$ \\
    MSE$_{\mathrm{PCA50}}$ $\downarrow$ & $\mathbf{8.02 \pm 0.33}$ & $8.07 \pm 0.32$ & $12.61 \pm 0.59$ & $8.04 \pm 0.33$ \\
    $W_1$ $\downarrow$ & $33.77 \pm 1.21$ & $\mathbf{33.76 \pm 1.18}$ & $46.77 \pm 1.23$ & $33.80 \pm 1.15$ \\
    $W_2$ $\downarrow$ & $578.1 \pm 42.4$ & $\mathbf{576.9 \pm 41.3}$ & $1108.5 \pm 58.9$ & $577.1 \pm 40.2$ \\
    Energy $\downarrow$ & $\mathbf{9.73 \pm 0.91}$ & $9.80 \pm 0.92$ & $14.89 \pm 0.97$ & $9.74 \pm 0.91$ \\
    \bottomrule
  \end{tabular}
\end{table}

\subsection{Sensitivity Studies}
\label{sec:sensitivity}

\begin{figure*}[p]
  \centering
  \begin{minipage}{0.49\textwidth}\centering\includegraphics[width=\textwidth]{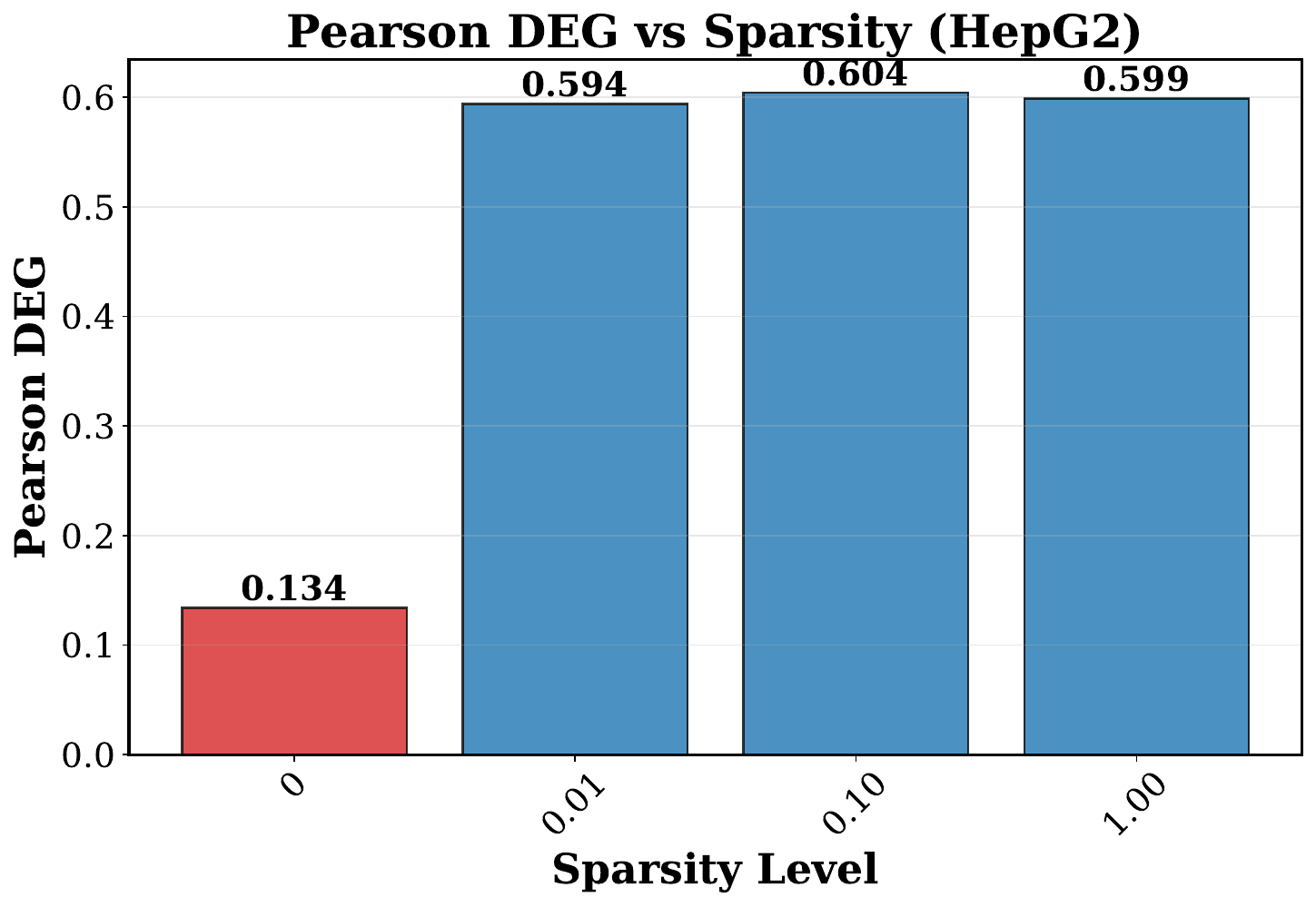}\end{minipage}
  \begin{minipage}{0.49\textwidth}\centering\includegraphics[width=\textwidth]{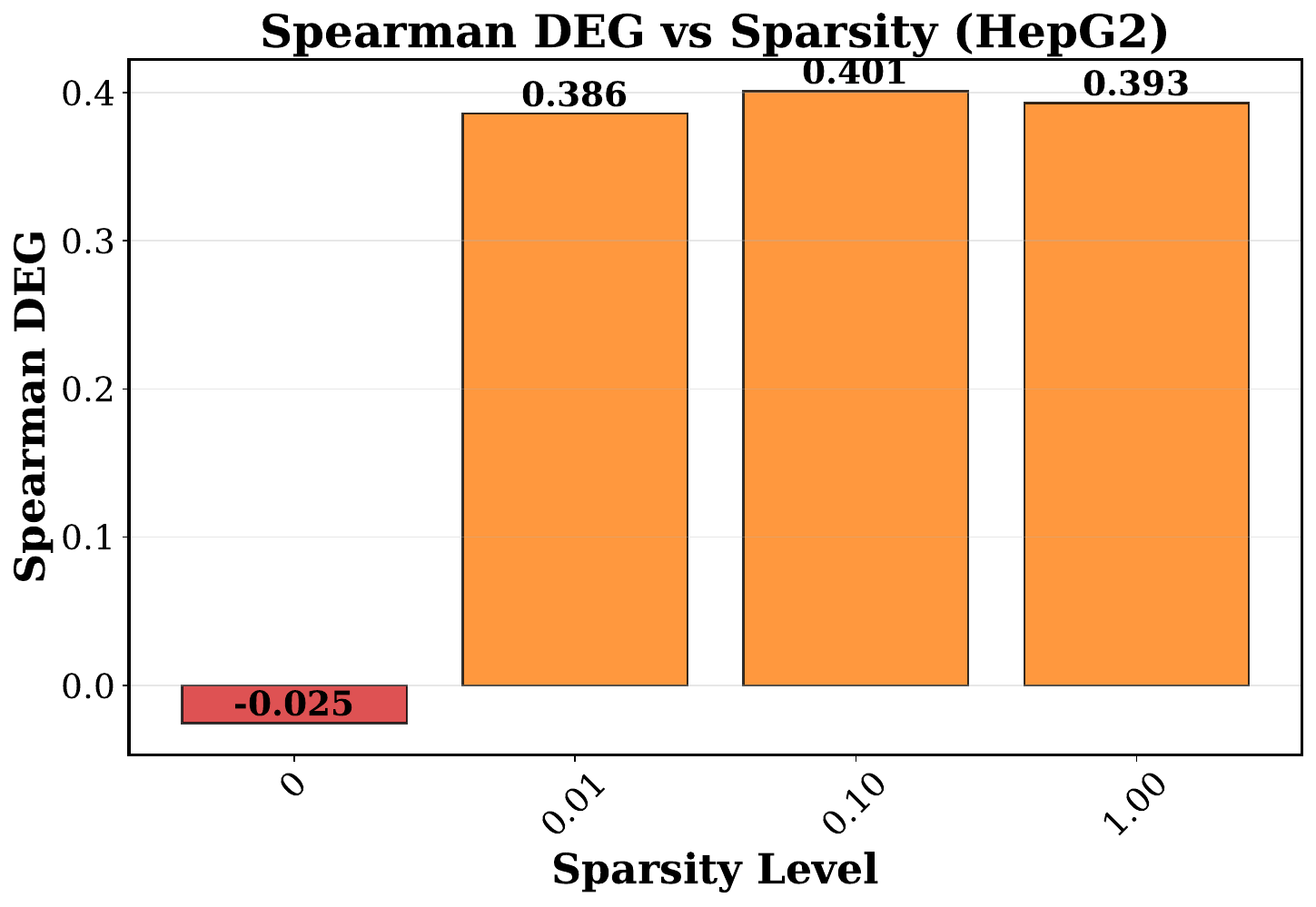}\end{minipage}\\[0.3cm]
  \begin{minipage}{0.49\textwidth}\centering\includegraphics[width=\textwidth]{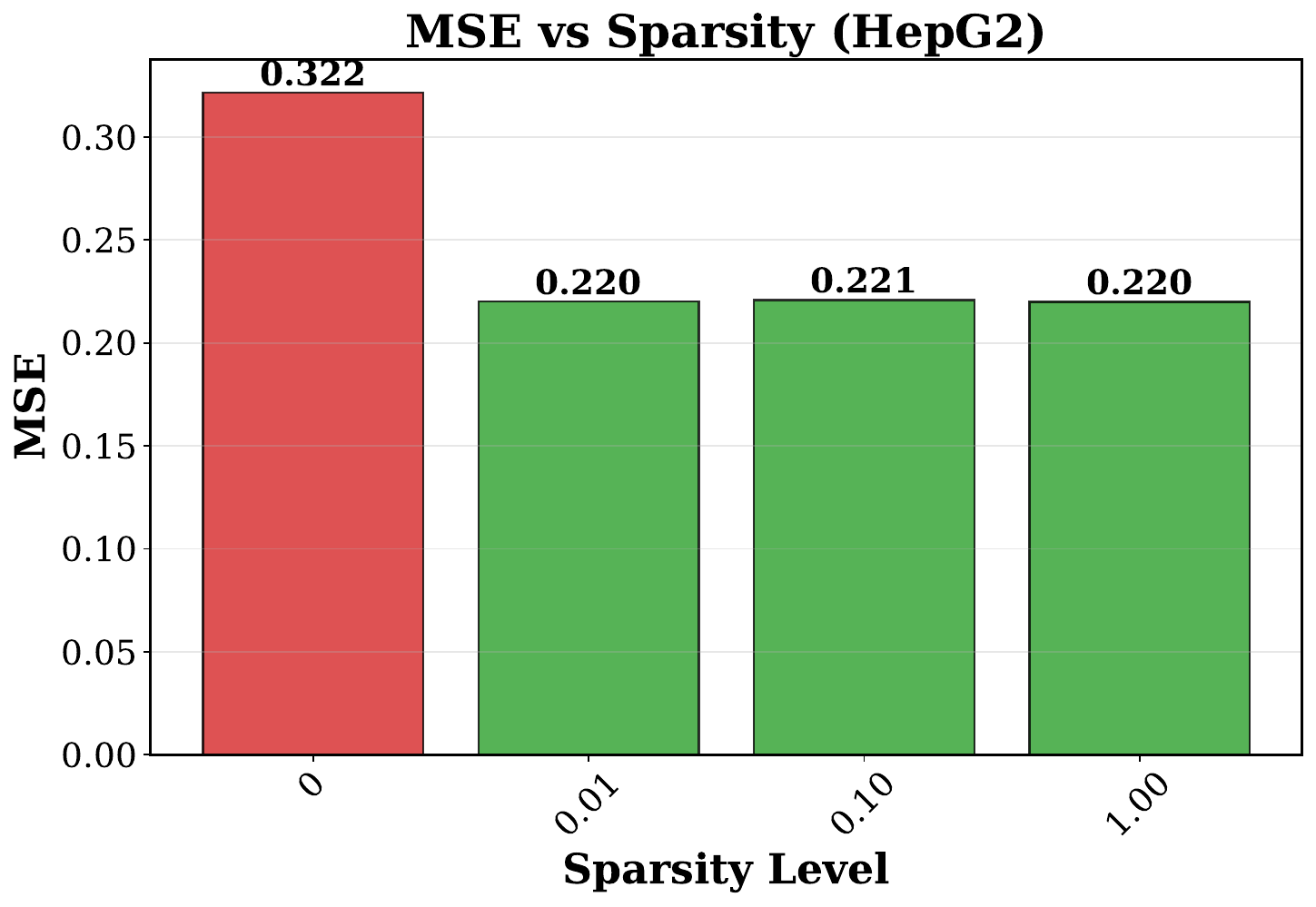}\end{minipage}
  \begin{minipage}{0.49\textwidth}\centering\includegraphics[width=\textwidth]{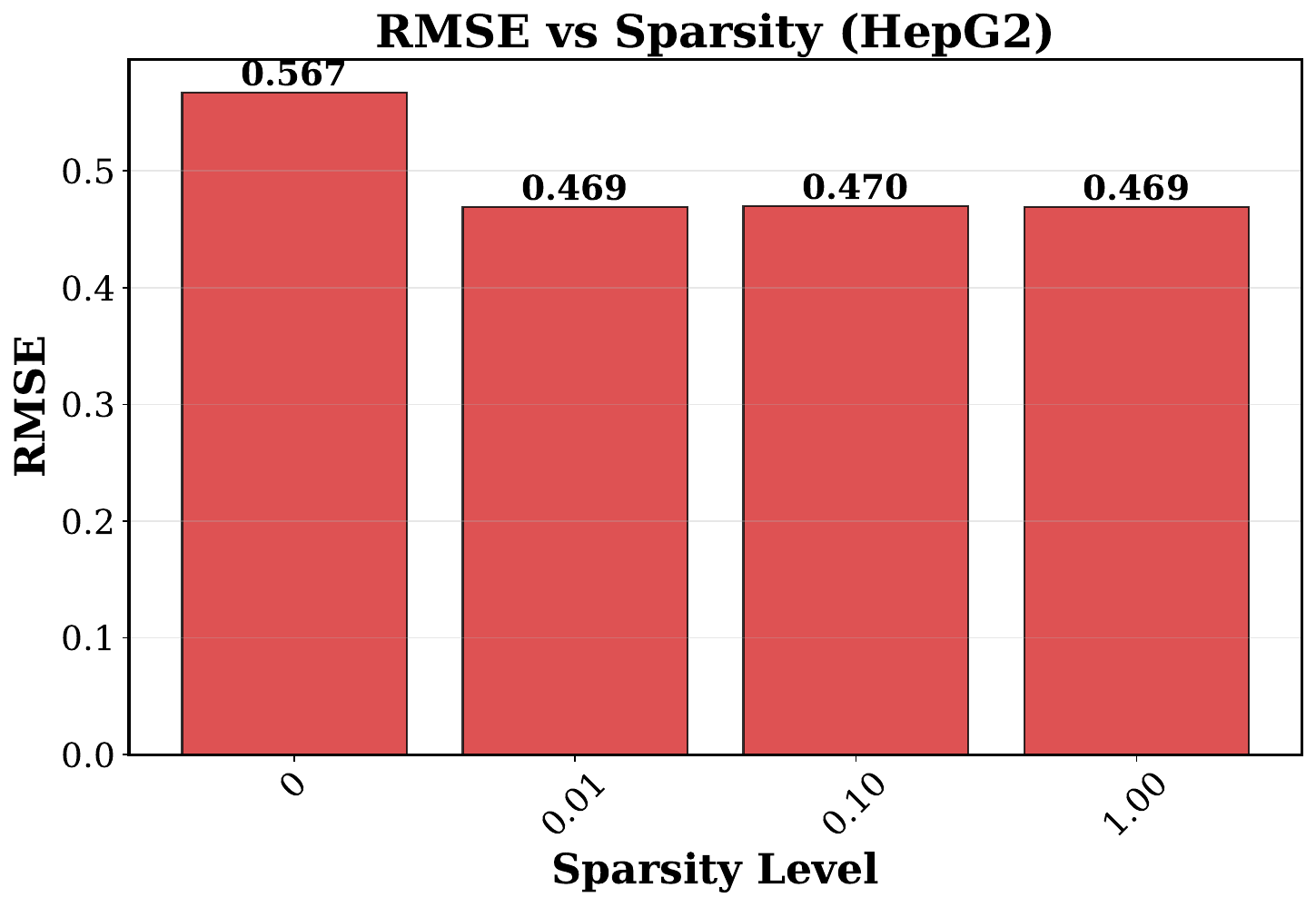}\end{minipage}
  \caption{\textbf{Sparsity regularization sensitivity (gene-level and reconstruction metrics).} HepG2, $K = 32$.}
  \label{fig:sensitivity_sparsity_main}
\end{figure*}

\begin{figure*}[p]
  \centering
  \begin{minipage}{0.49\textwidth}\centering\includegraphics[width=\textwidth]{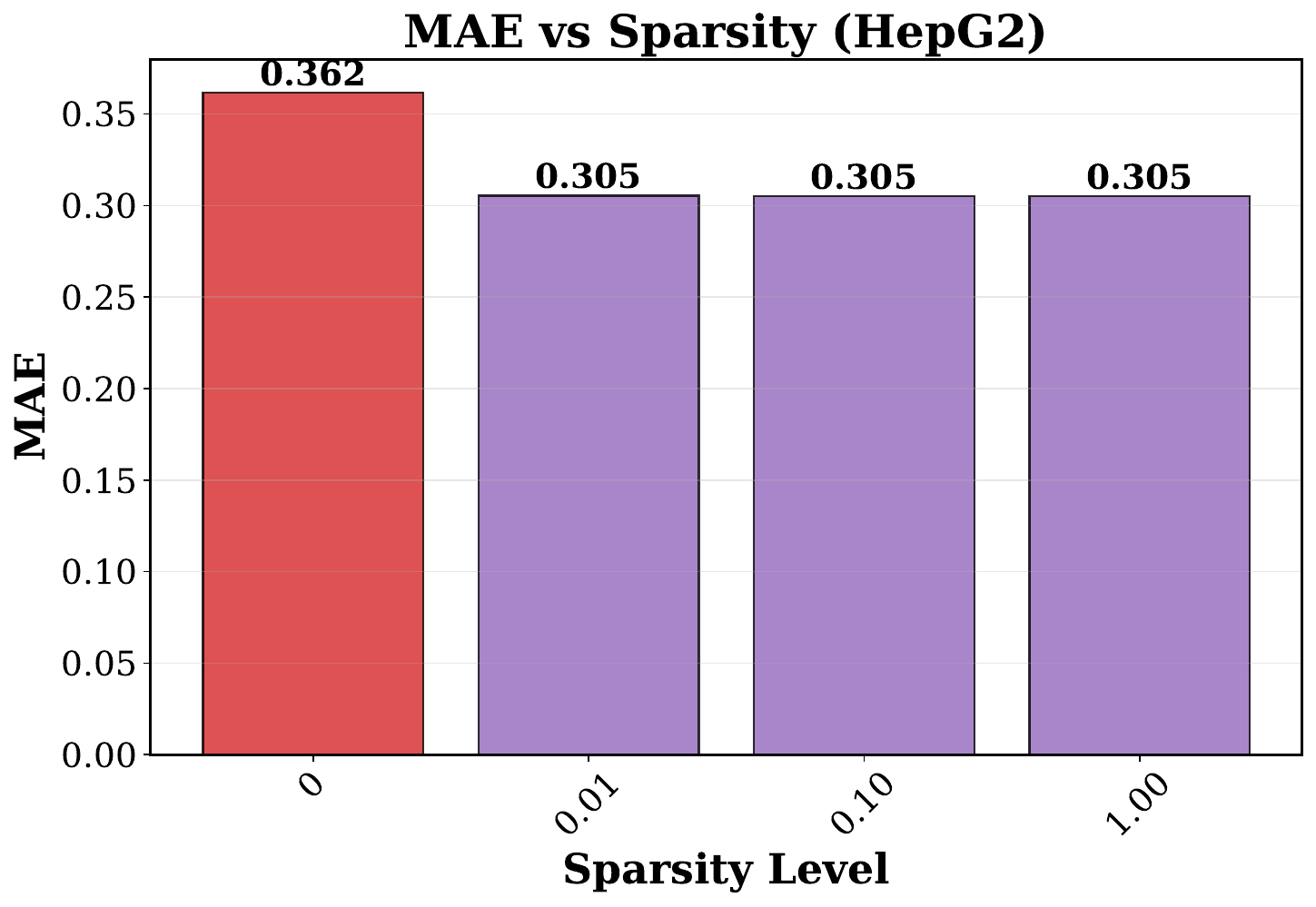}\end{minipage}
  \begin{minipage}{0.49\textwidth}\centering\includegraphics[width=\textwidth]{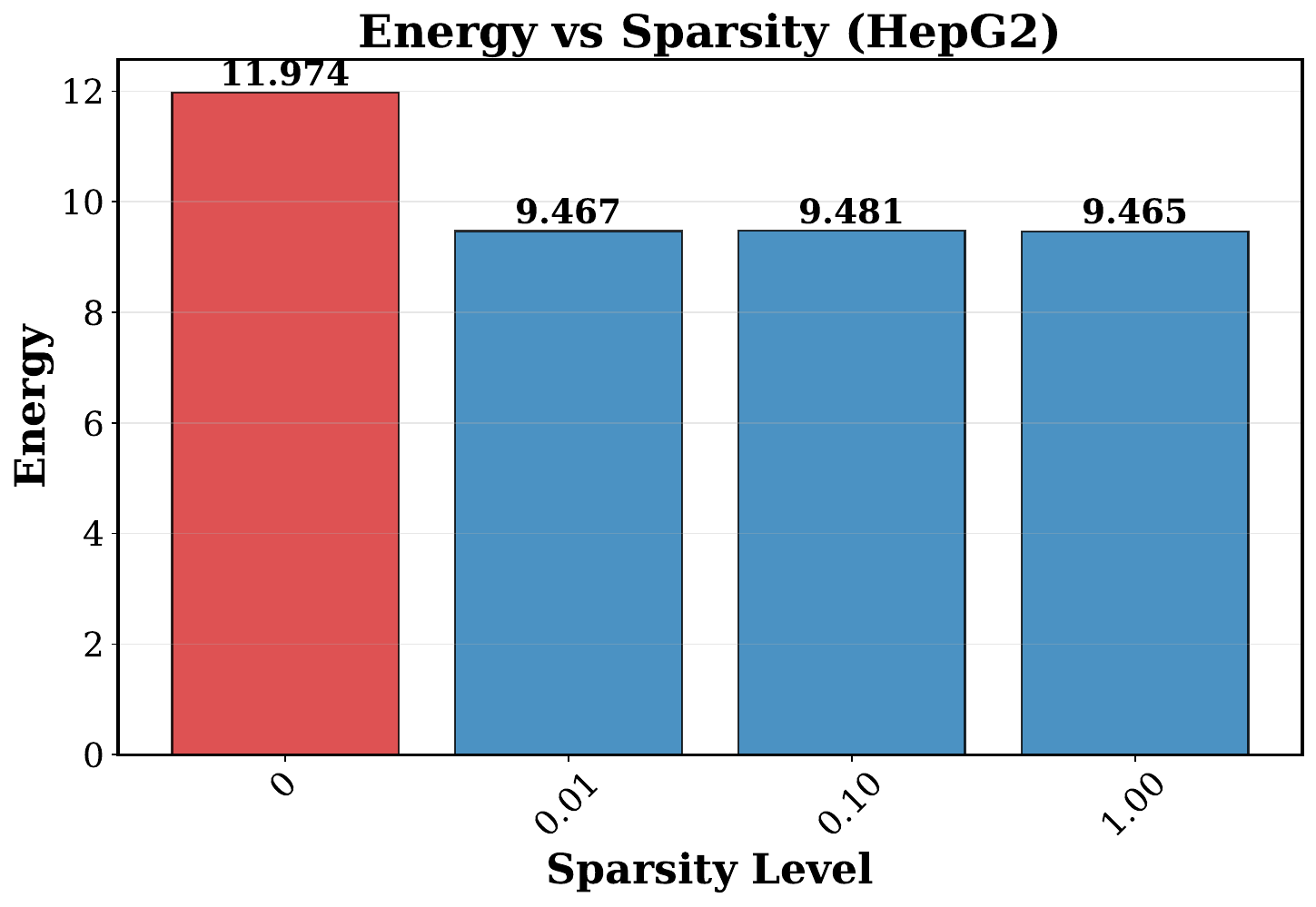}\end{minipage}\\[0.3cm]
  \begin{minipage}{0.49\textwidth}\centering\includegraphics[width=\textwidth]{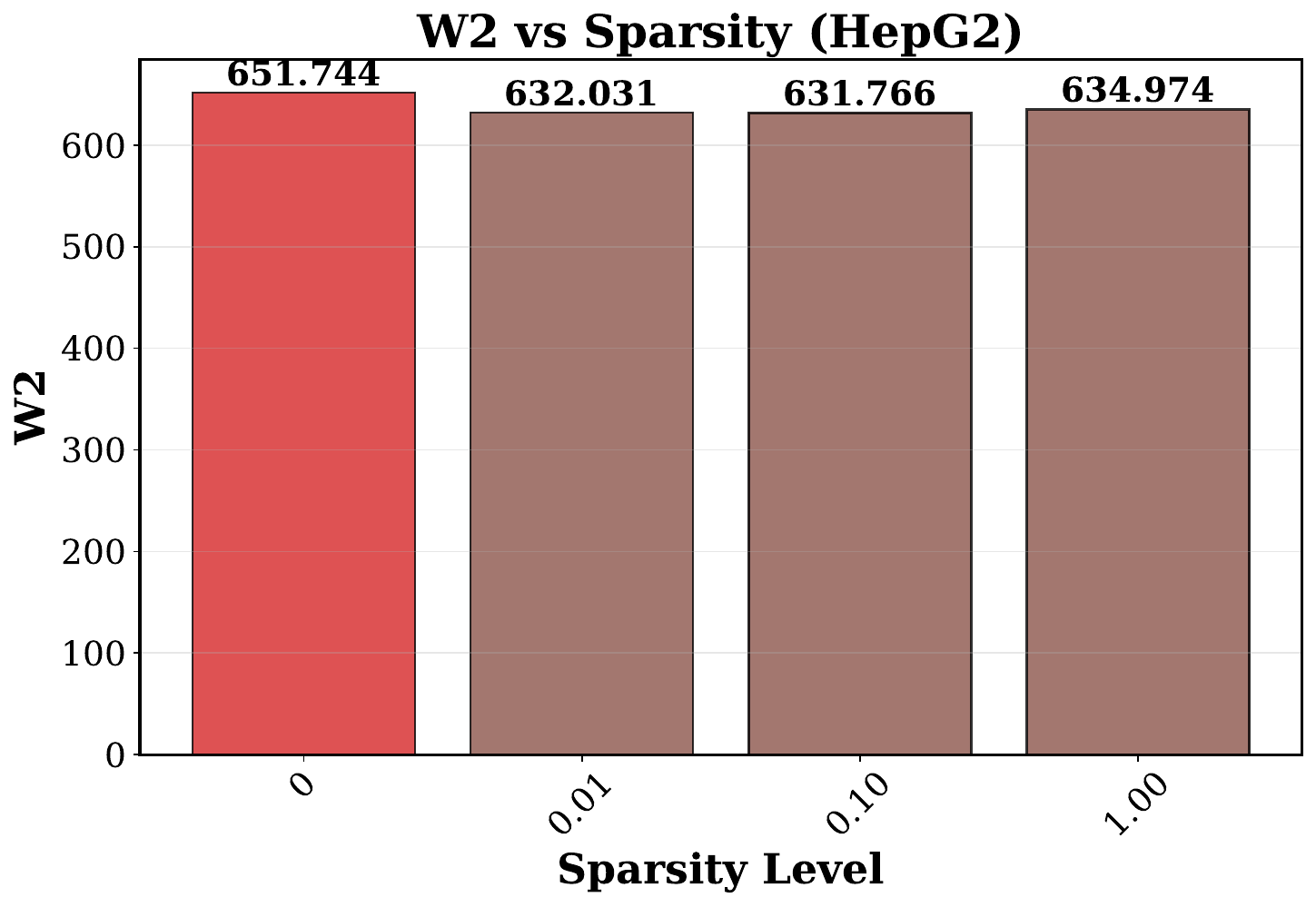}\end{minipage}
  \begin{minipage}{0.49\textwidth}\centering\includegraphics[width=\textwidth]{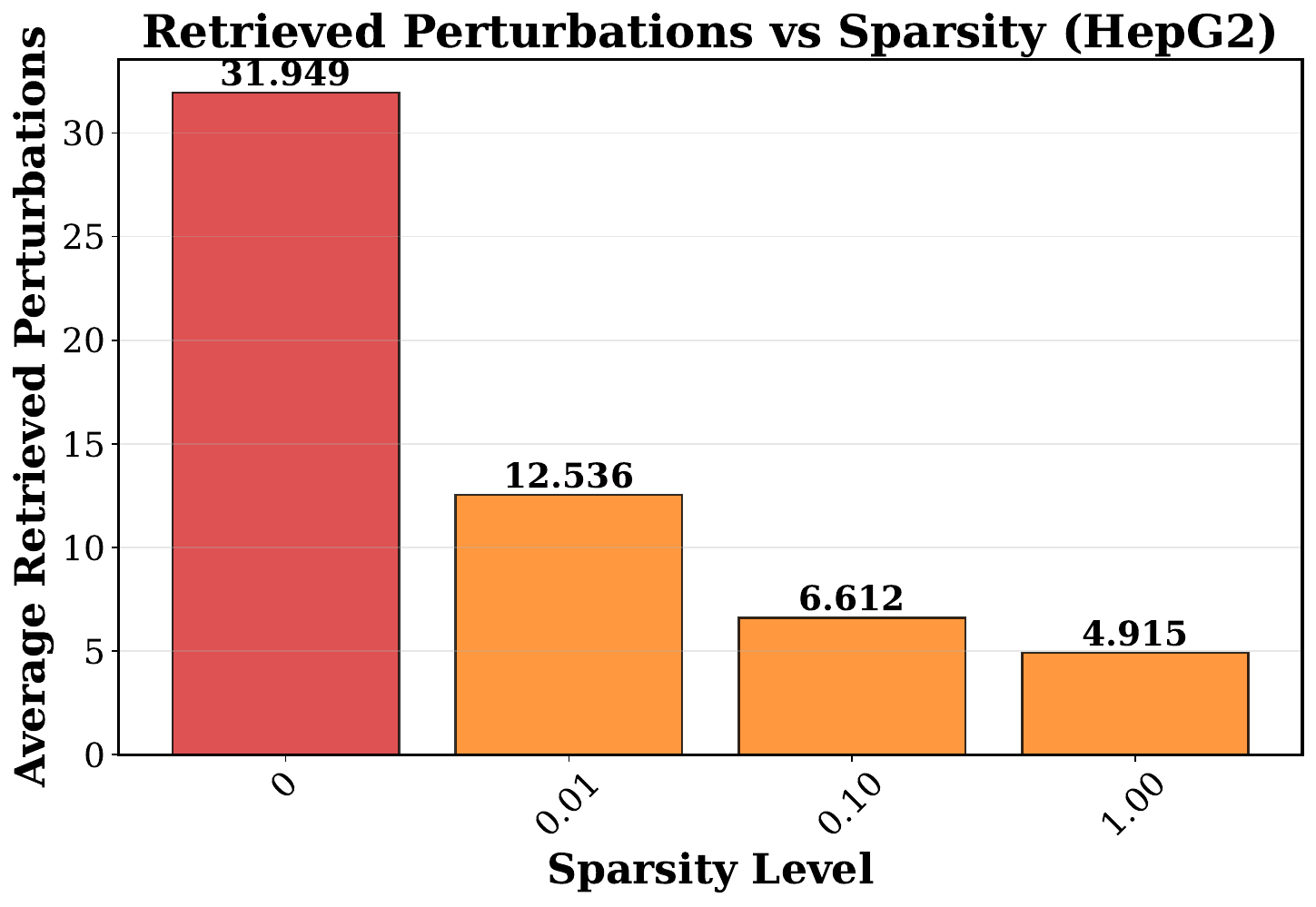}\end{minipage}
  \caption{\textbf{Sparsity regularization sensitivity (distributional metrics and retrieval count).} HepG2, $K = 32$.}
  \label{fig:sensitivity_sparsity_dist}
\end{figure*}

\textbf{Sparsity regularization.}
Figures~\ref{fig:sensitivity_sparsity_main} and \ref{fig:sensitivity_sparsity_dist} analyze $\lambda_{\mathrm{sparse}} \in \{0, 0.01, 0.10, 1.00\}$ on HepG2 ($K = 32$).
Without sparsity ($\lambda = 0$), the model retrieves $\approx$32 candidates on average, yielding Pearson DEG 0.134 (vs.\ 0.604 with $\lambda = 0.1$), Spearman $-0.025$, and RMSE 0.567.
With any non-zero $\lambda$, performance stabilizes: Pearson 0.594--0.604, Spearman 0.386--0.401, RMSE $\approx$0.470, average selected perturbations 4.9--12.5.

\begin{figure*}[p]
  \centering
  \begin{minipage}{0.49\textwidth}\centering\includegraphics[width=\textwidth]{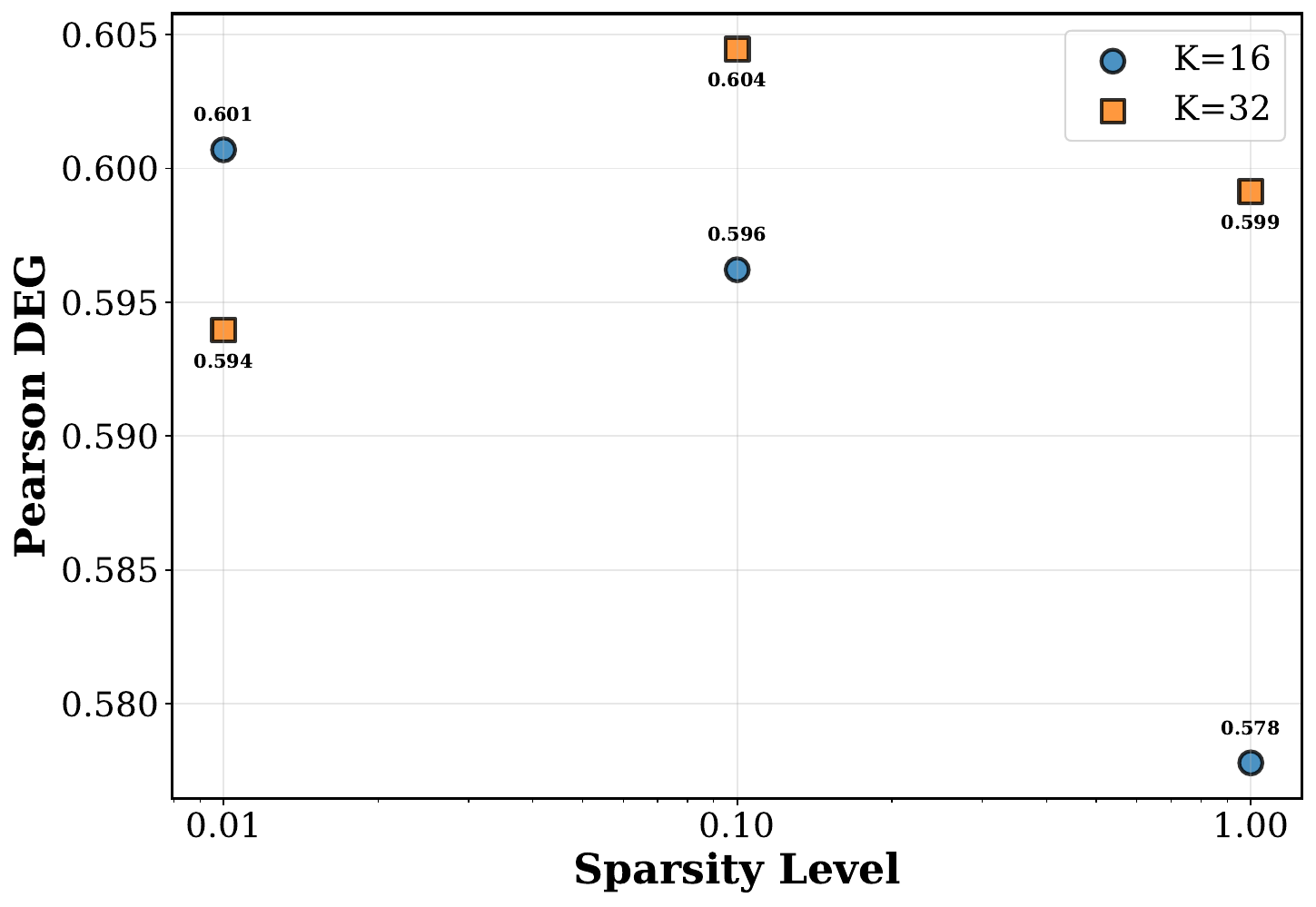}\end{minipage}
  \begin{minipage}{0.49\textwidth}\centering\includegraphics[width=\textwidth]{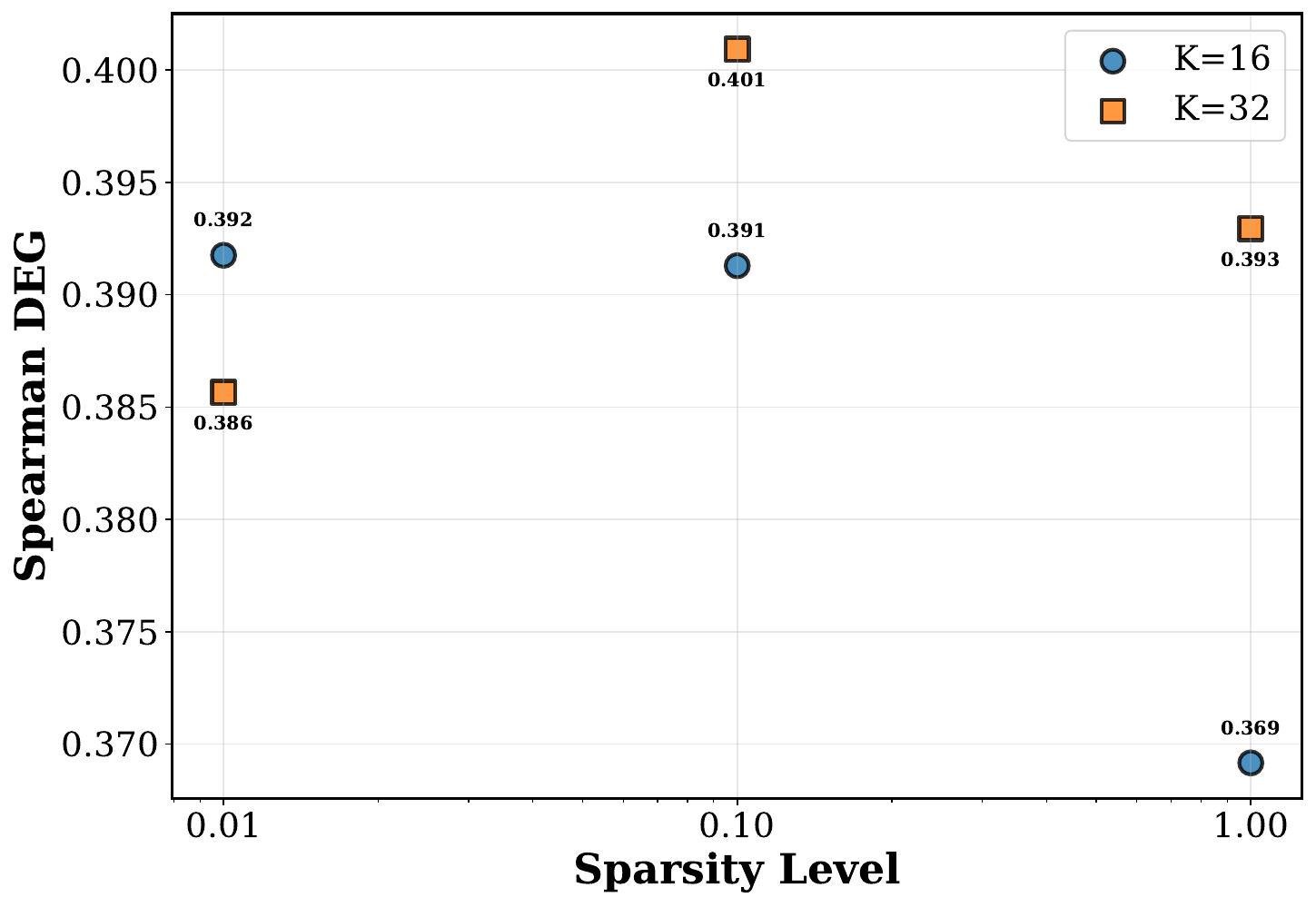}\end{minipage}\\[0.3cm]
  \begin{minipage}{0.49\textwidth}\centering\includegraphics[width=\textwidth]{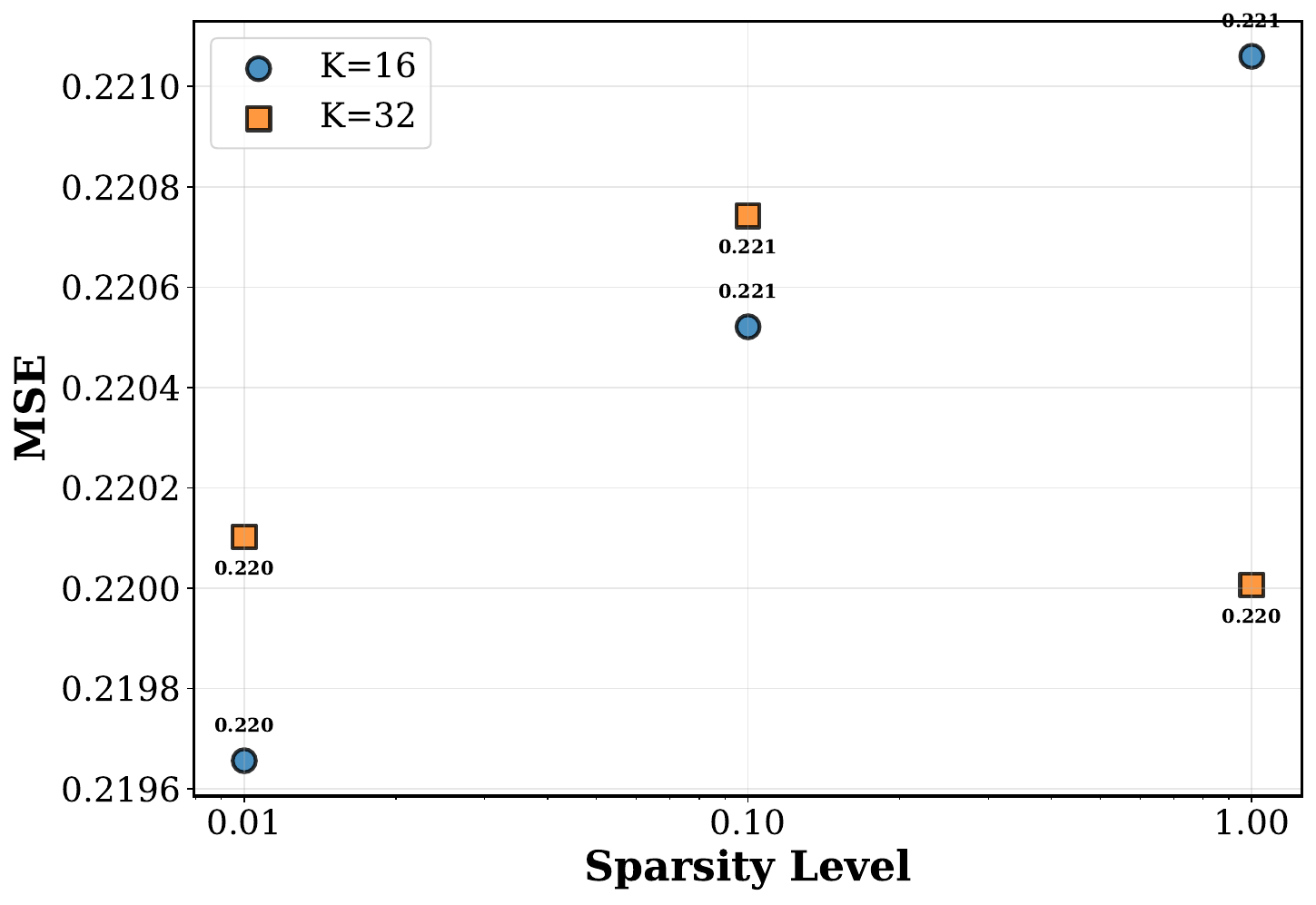}\end{minipage}
  \begin{minipage}{0.49\textwidth}\centering\includegraphics[width=\textwidth]{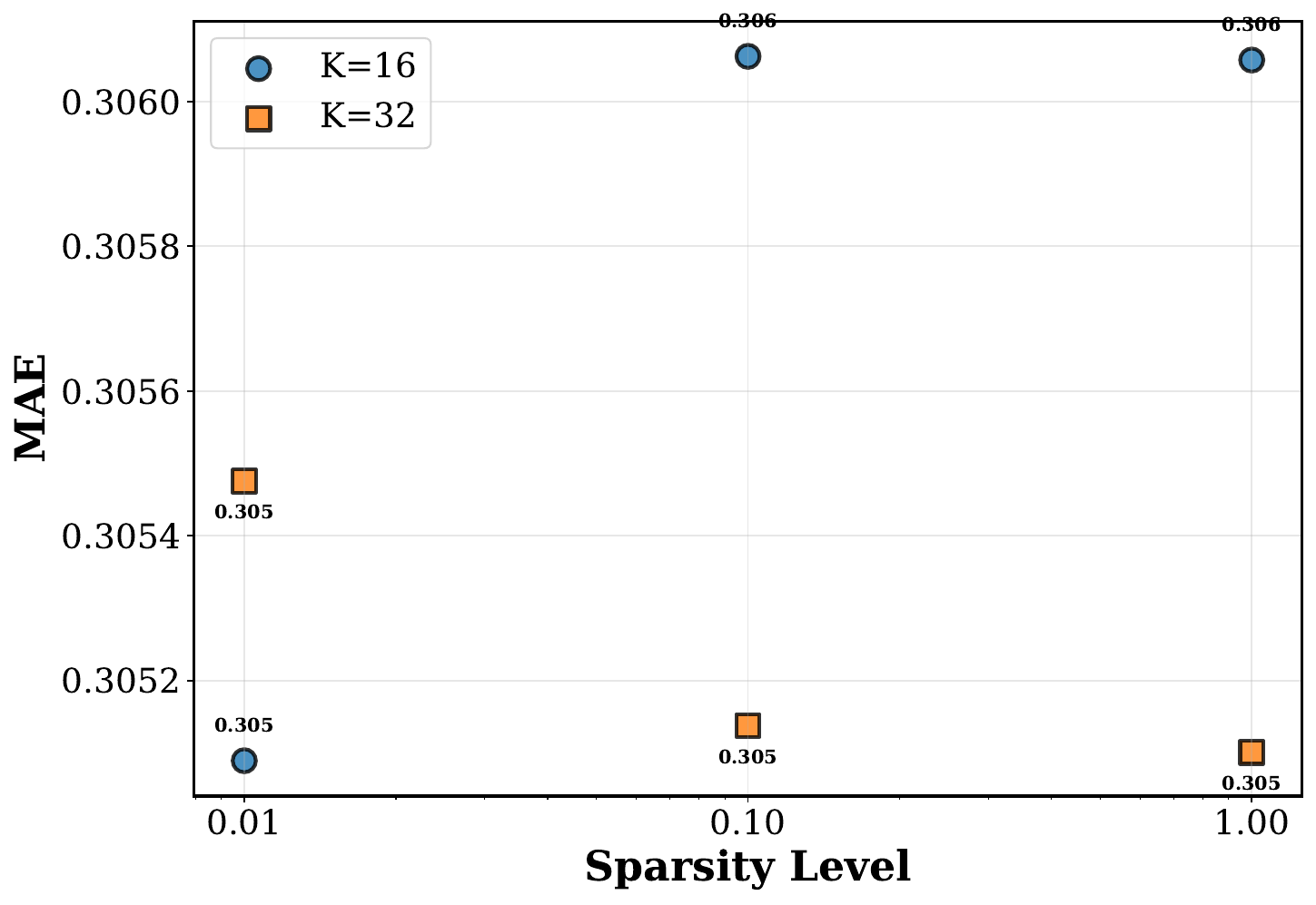}\end{minipage}
  \caption{\textbf{$K = 16$ vs.\ $K = 32$ (gene-level and reconstruction metrics).} HepG2.}
  \label{fig:sensitivity_k_comparison_main}
\end{figure*}

\begin{figure*}[p]
  \centering
  \begin{minipage}{0.49\textwidth}\centering\includegraphics[width=\textwidth]{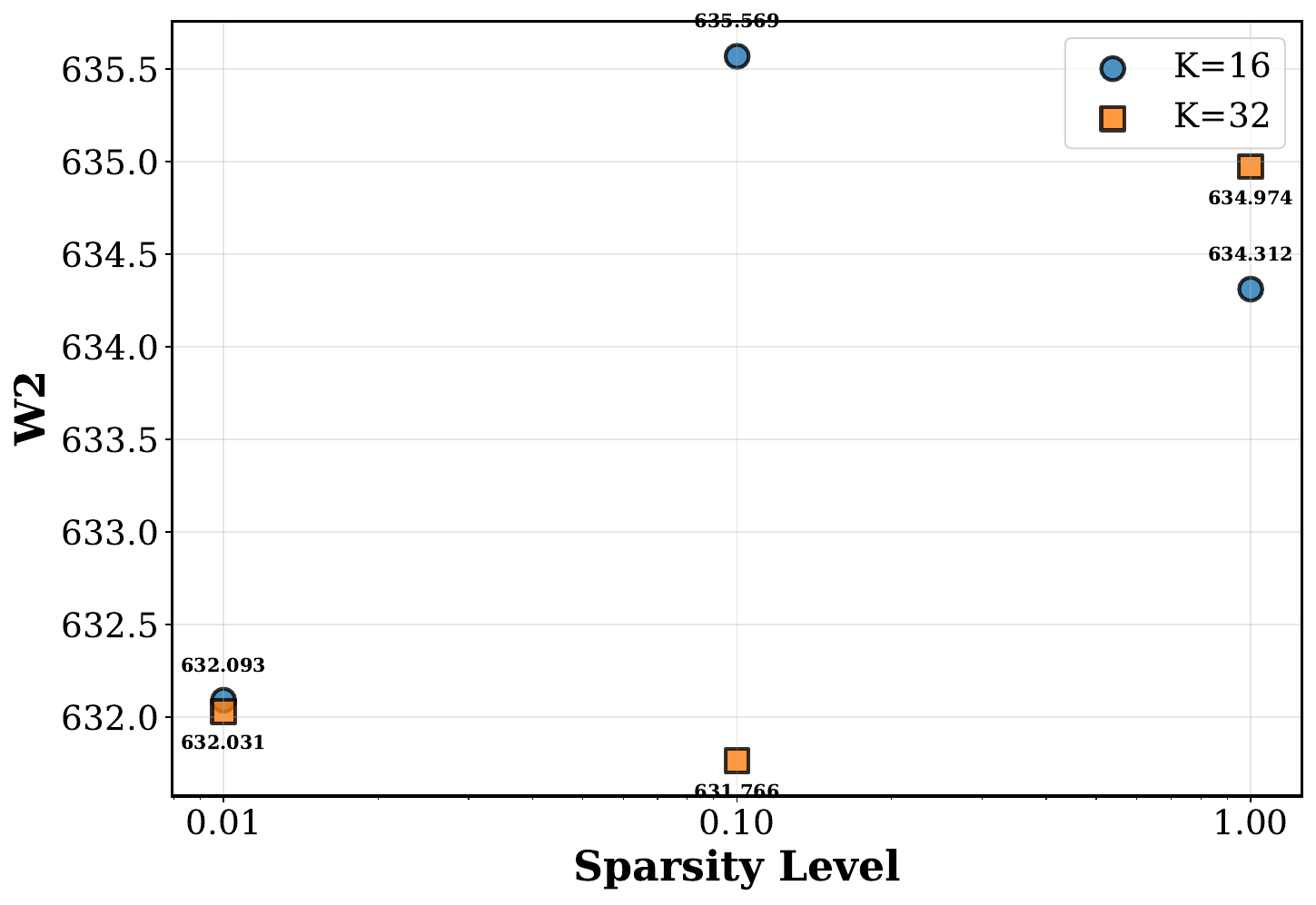}\end{minipage}
  \begin{minipage}{0.49\textwidth}\centering\includegraphics[width=\textwidth]{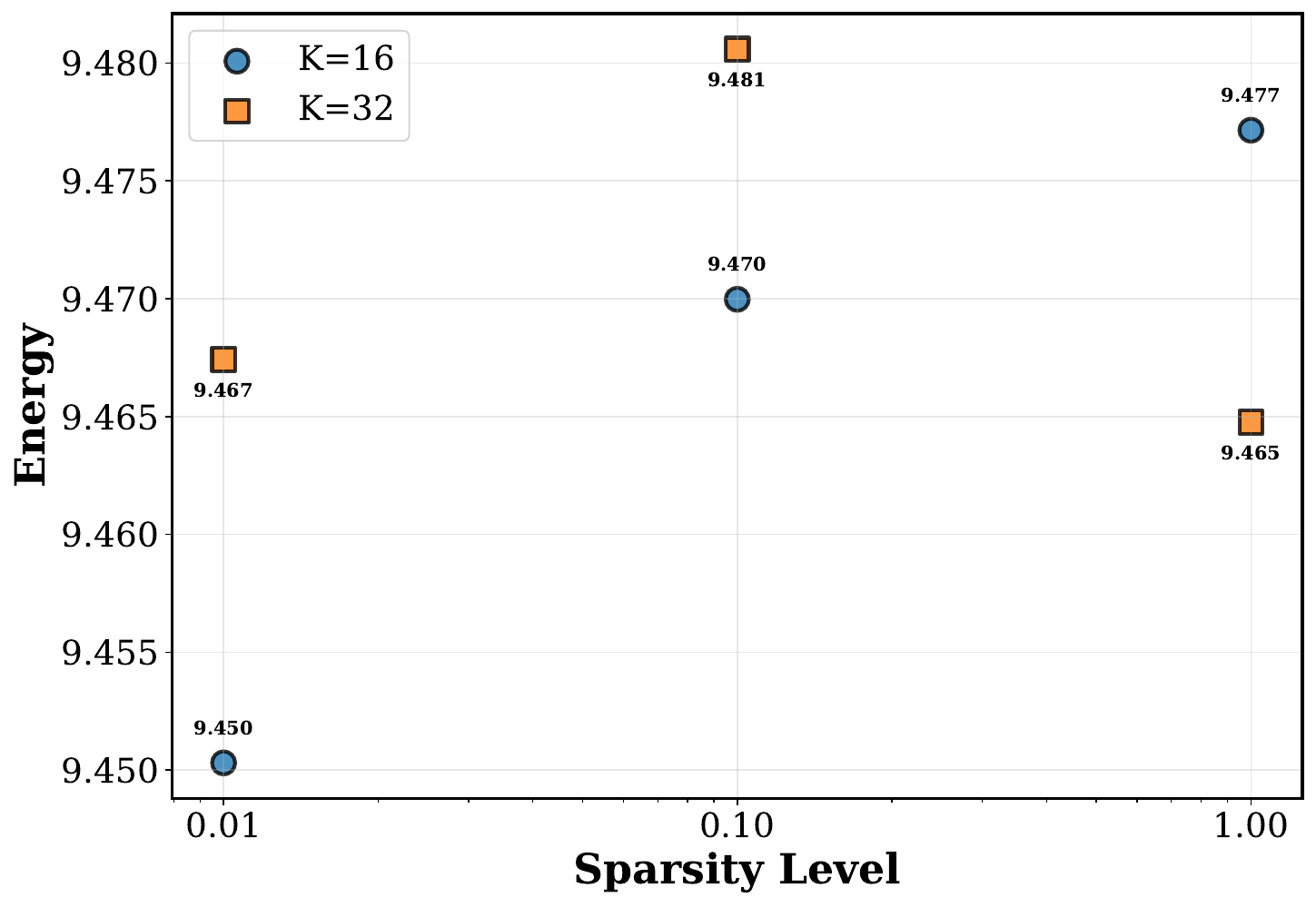}\end{minipage}
  \caption{\textbf{$K = 16$ vs.\ $K = 32$ (distributional metrics).} HepG2.}
  \label{fig:sensitivity_k_comparison_dist}
\end{figure*}

\textbf{Retrieval size $K$.}
Figures~\ref{fig:sensitivity_k_comparison_main} and \ref{fig:sensitivity_k_comparison_dist} compare $K \in \{16, 32\}$ across sparsity levels on HepG2.
Both settings yield comparable performance across all sparsity levels, demonstrating that PT-RAG is robust to $K$ within this range.

\begin{figure*}[p]
  \centering
  \begin{minipage}{0.49\textwidth}\centering\includegraphics[width=\textwidth]{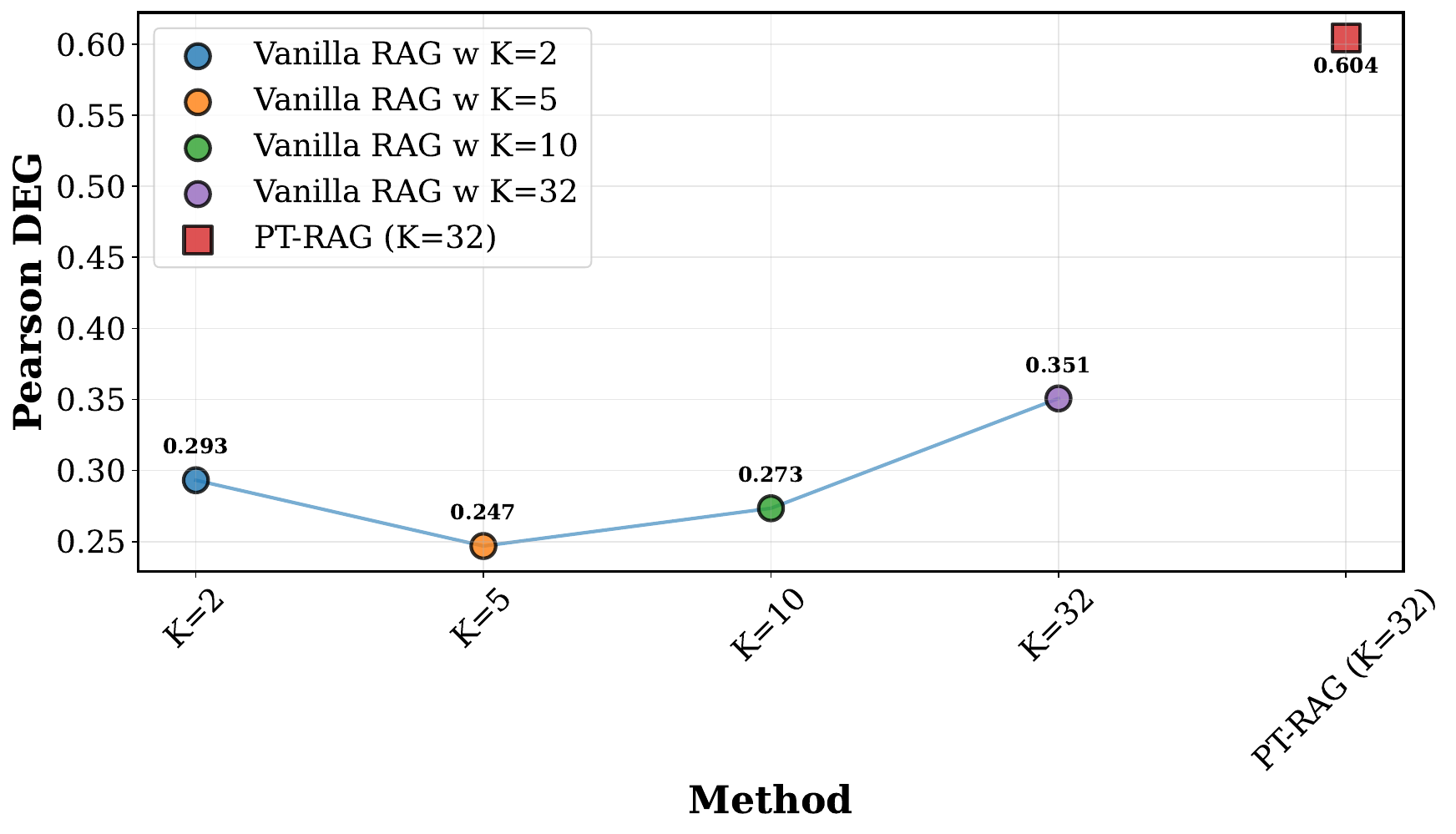}\end{minipage}
  \begin{minipage}{0.49\textwidth}\centering\includegraphics[width=\textwidth]{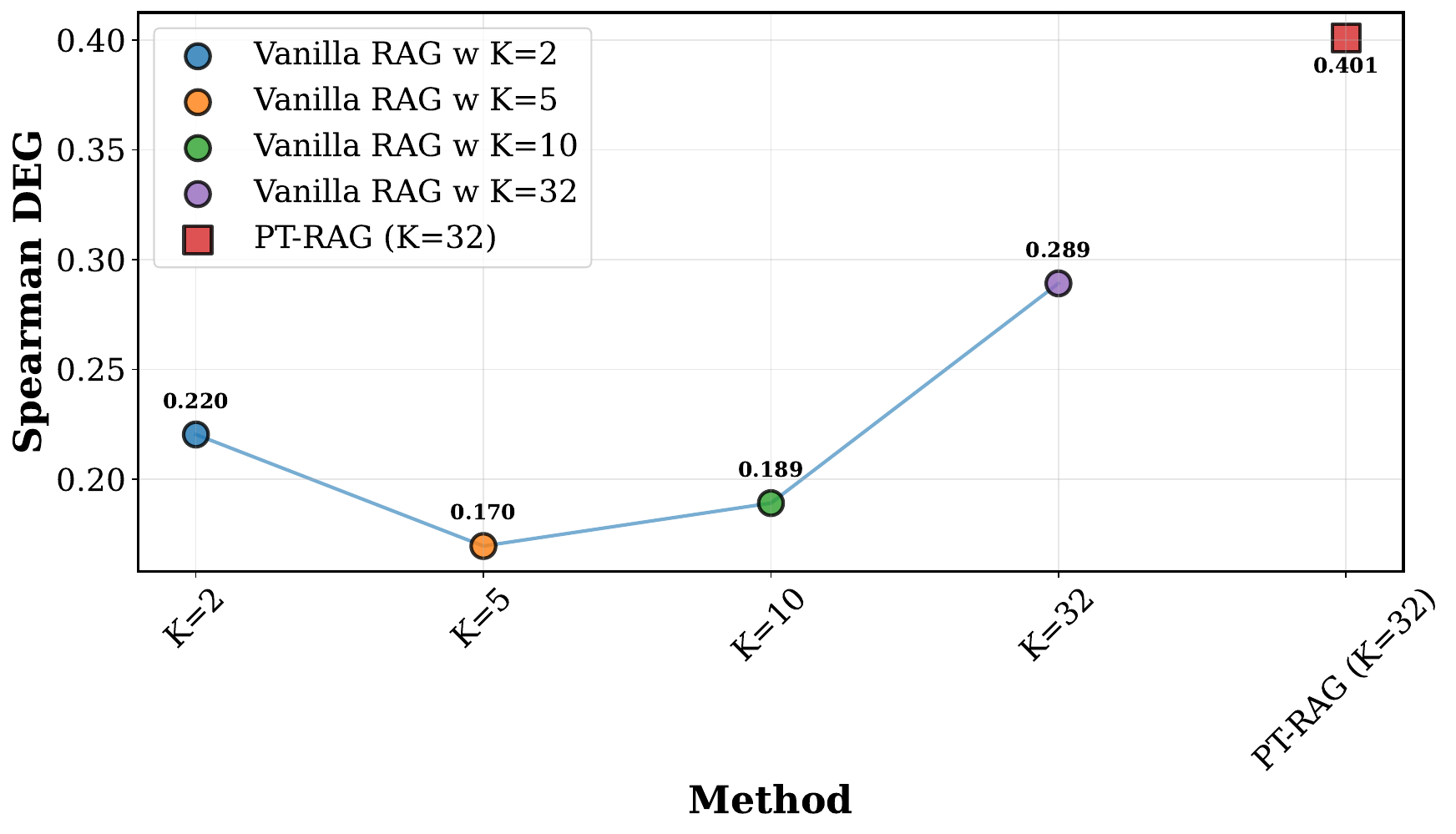}\end{minipage}\\[0.3cm]
  \begin{minipage}{0.49\textwidth}\centering\includegraphics[width=\textwidth]{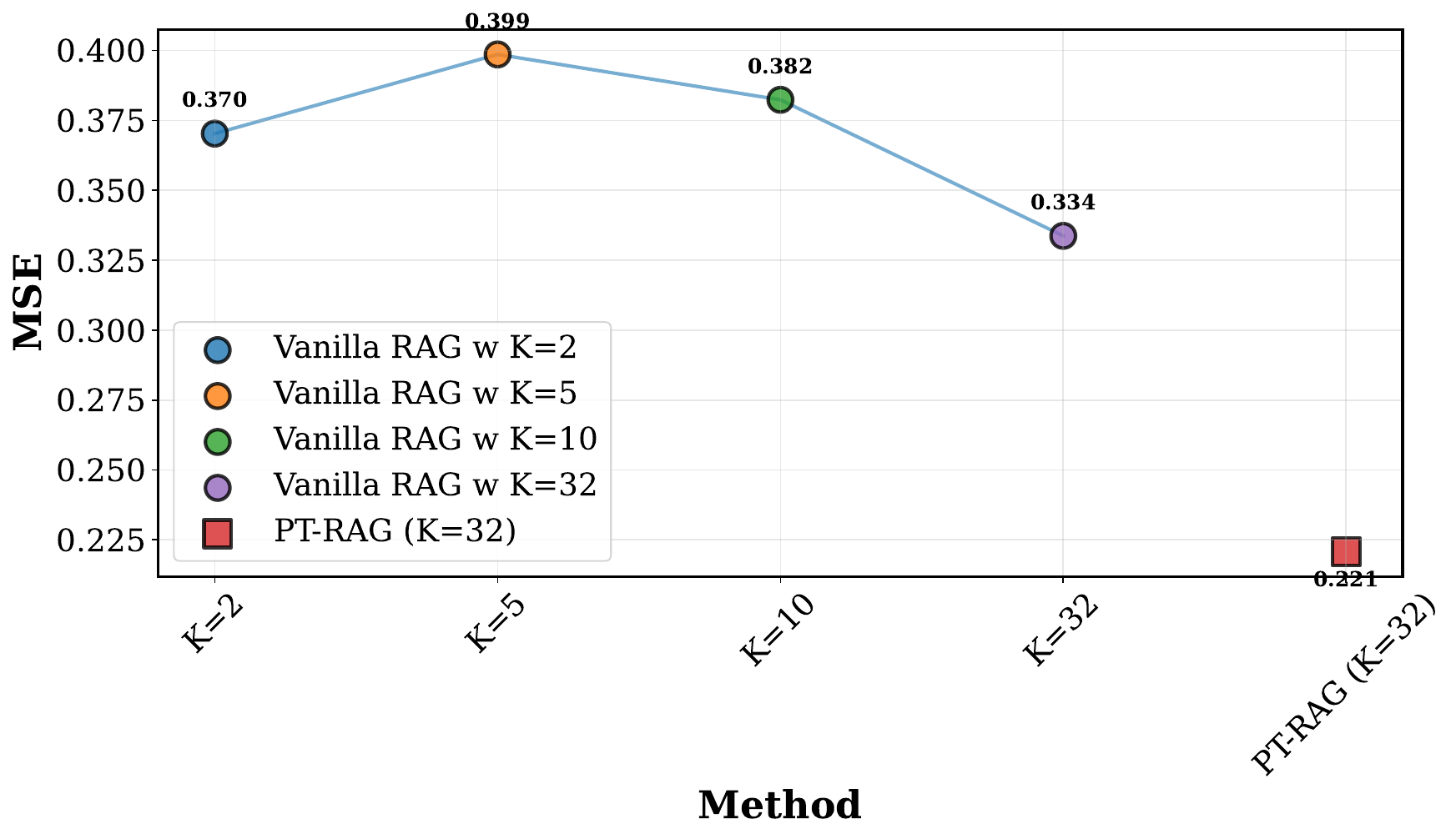}\end{minipage}
  \begin{minipage}{0.49\textwidth}\centering\includegraphics[width=\textwidth]{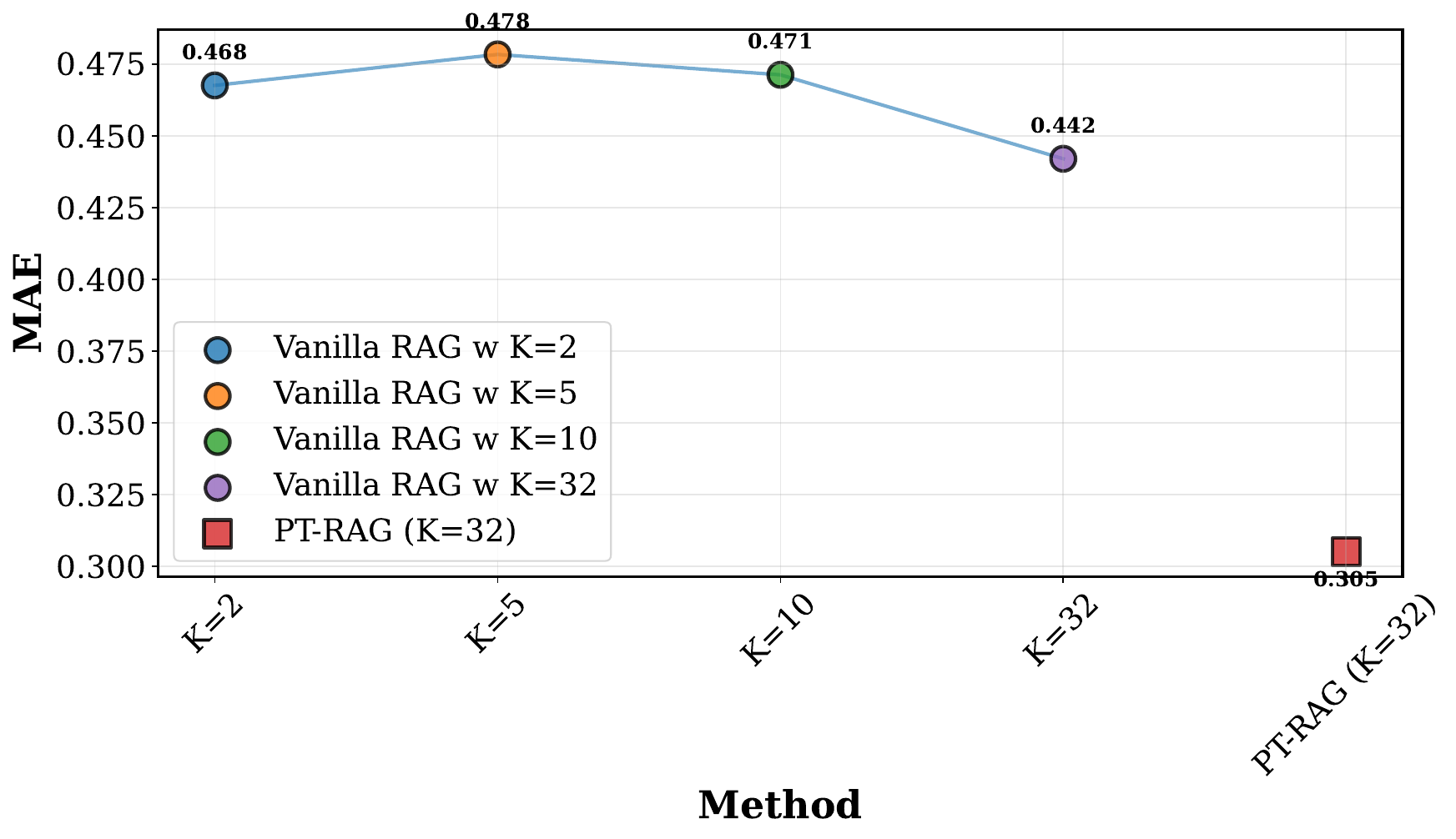}\end{minipage}
  \caption{\textbf{Naïve RAG $K \in \{2, 5, 10, 32\}$ vs.\ PT-RAG ($K=32$).} Gene-level and reconstruction. HepG2.}
  \label{fig:naïve_rag_k_ablation_core}
\end{figure*}

\begin{figure*}[p]
  \centering
  \begin{minipage}{0.49\textwidth}\centering\includegraphics[width=\textwidth]{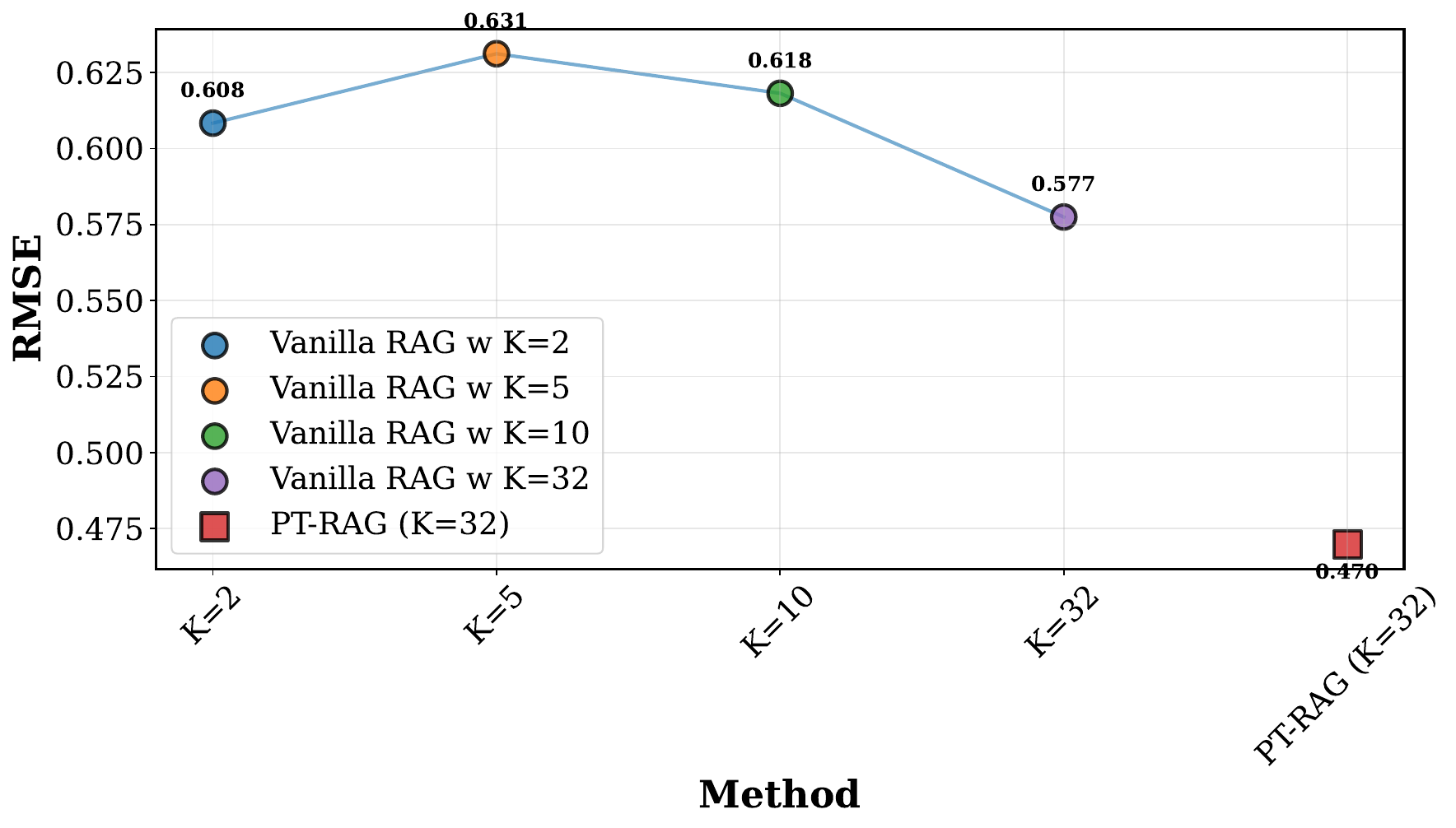}\end{minipage}
  \begin{minipage}{0.49\textwidth}\centering\includegraphics[width=\textwidth]{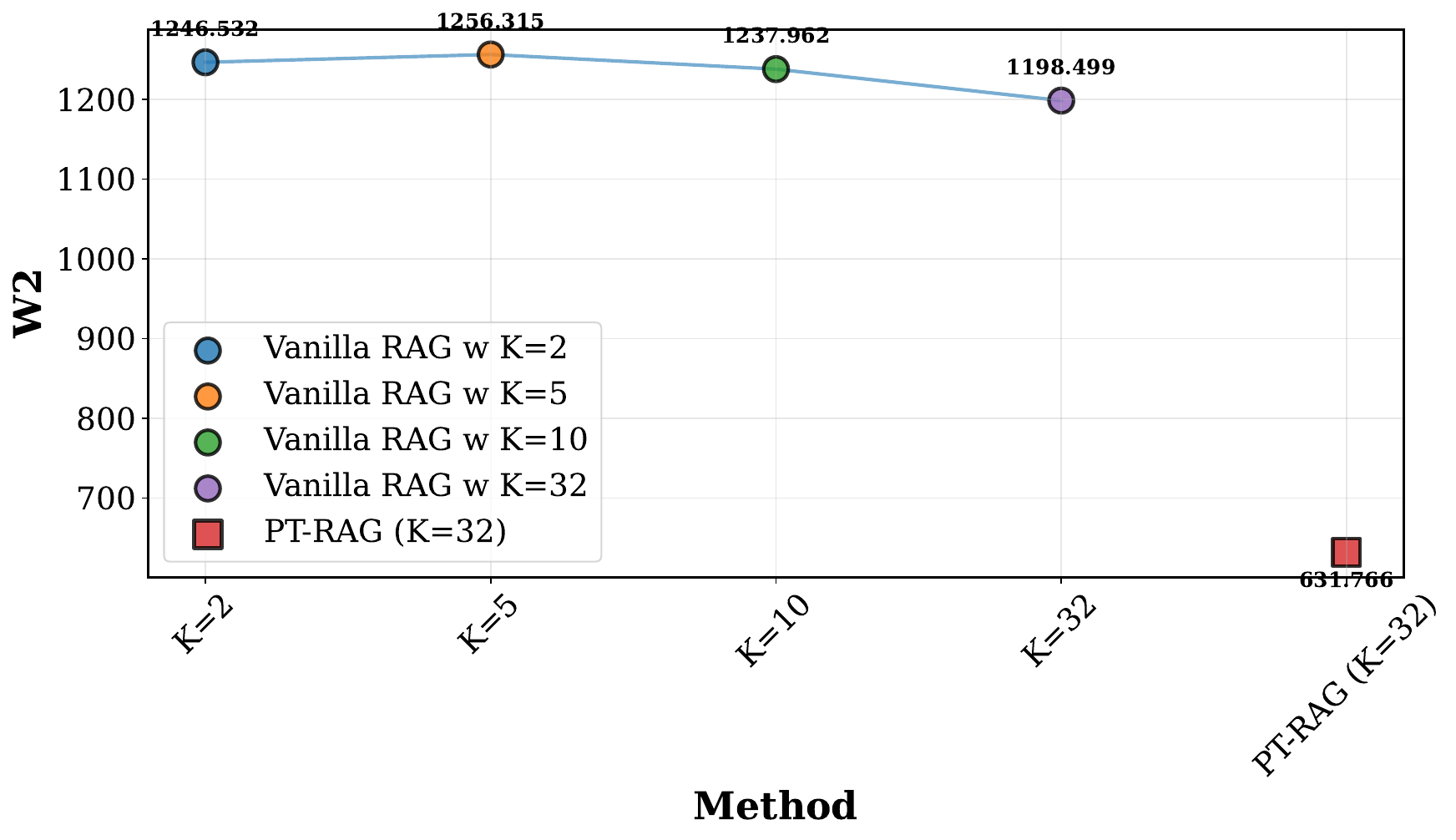}\end{minipage}\\[0.3cm]
  \begin{minipage}{0.49\textwidth}\centering\includegraphics[width=\textwidth]{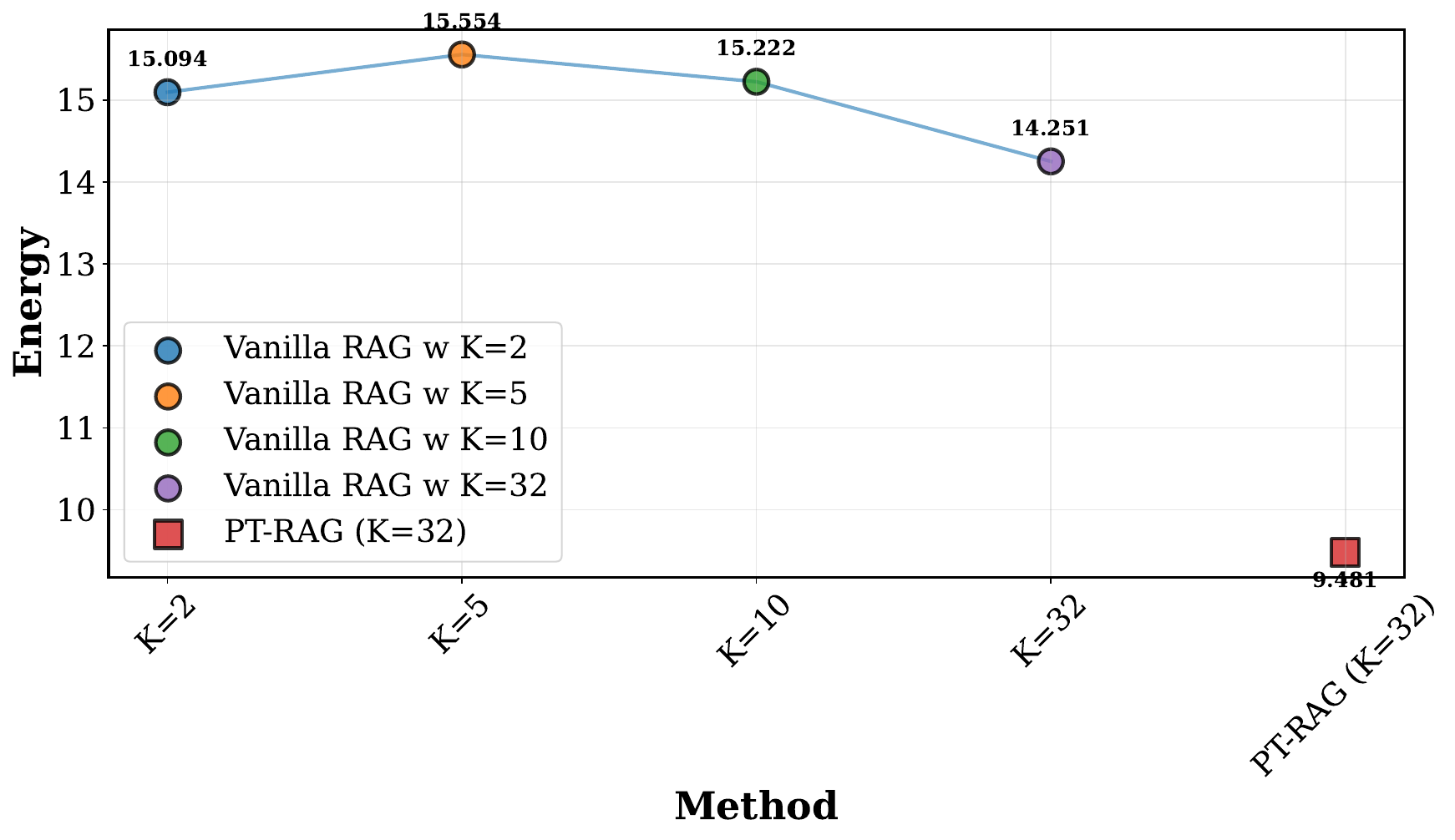}\end{minipage}
  \caption{\textbf{Naïve RAG $K \in \{2, 5, 10, 32\}$ vs.\ PT-RAG ($K=32$).} Distributional metrics. HepG2.}
  \label{fig:naïve_rag_k_ablation_dist}
\end{figure*}

\textbf{Naïve RAG retrieval size ablation.}
Figures~\ref{fig:naïve_rag_k_ablation_core} and \ref{fig:naïve_rag_k_ablation_dist} evaluate Naïve RAG with $K \in \{2, 5, 10, 32\}$ vs.\ PT-RAG ($K = 32$) on HepG2.
At $K = 32$, Naïve RAG reaches Pearson 0.351 vs.\ PT-RAG's 0.604 (gap: 42\%).
Spearman is non-monotonic with $K$ (0.220 at $K\!=\!2$, dipping to 0.170 at $K\!=\!5$, recovering to 0.289 at $K\!=\!32$), and RMSE degrades from $K\!=\!2$ to $K\!=\!5$ before recovering.
This non-monotonic behavior confirms that the failure is not due to insufficient retrieval breadth but to the lack of differentiable, cell-type-aware selection.

\subsection{Additional Controls}
\label{sec:additional_controls}

This section reports controls that isolate where the gains of PT-RAG come from. Unless stated otherwise, they are run in the cross-perturbation setting on a single fold (3{,}452 test cell-type--perturbation pairs) with a PT-RAG model using $K = 32$. Since they use one fold and a different $K$, their values should be compared with each other and with the STATE+GenePT row of the same table, not with the three-fold averages of Table~\ref{tab:crosspert_results}.

\begin{table}[h]
\centering
\caption{\textbf{Additional controls, cross-perturbation (one fold, $K = 32$).} See the text for the description of each row.}
\label{tab:additional_controls}
\small
\setlength{\tabcolsep}{3pt}
\begin{tabular}{lcccccc}
\toprule
\textbf{Model} & \textbf{Pearson} $\uparrow$ & \textbf{Spearman} $\uparrow$ & \textbf{MSE$_{\mathrm{PCA50}}$} $\downarrow$ & $W_1$ $\downarrow$ & $W_2$ $\downarrow$ & \textbf{Energy} $\downarrow$ \\
\midrule
STATE+GenePT (no retrieval) & 0.6409 & 0.4314 & 1.3056 & 0.8166 & 0.7173 & 0.3618 \\
\midrule
\multicolumn{7}{l}{\textit{Retrieval pool available at test time}} \\
PT-RAG, full pool (2{,}008) & 0.6242 & 0.4139 & 1.2556 & 0.8073 & 0.6939 & 0.3511 \\
PT-RAG, train+val pool (802) & 0.6237 & 0.4135 & 1.2552 & 0.8072 & 0.6937 & 0.3511 \\
PT-RAG, train-only pool (669) & 0.6237 & 0.4135 & 1.2551 & 0.8072 & 0.6936 & 0.3510 \\
\midrule
\multicolumn{7}{l}{\textit{Contribution of the selector and robustness}} \\
Frozen encoder + selector & 0.6432 & 0.4284 & 1.2880 & 0.8157 & 0.7104 & 0.3628 \\
PT-RAG, 60\% easy negatives & 0.6274 & 0.4171 & 1.2675 & 0.8096 & 0.6990 & 0.3541 \\
\bottomrule
\end{tabular}
\end{table}

\textbf{Retrieval pool.} In the cross-perturbation protocol, PT-RAG can retrieve from all perturbations at test time, whereas only the training perturbations are available during training (Section~\ref{sec:crossperturbation}). To check whether the larger test-time pool explains the gains, we evaluate the same trained model with the pool restricted to the train and validation perturbations, and to the training perturbations only. All metrics are essentially unchanged, so the improvements over STATE+GenePT on reconstruction and distributional metrics do not come from the larger pool.

\textbf{Contribution of the selector.} To separate the learned selection from the training of the perturbation encoder, we initialise the perturbation encoder from the STATE+GenePT model and keep it frozen, training only the scorer and the projector. This improves Pearson correlation, MSE$_{\mathrm{PCA50}}$, $W_1$ and $W_2$ over STATE+GenePT, but not Energy, and it does not reach the distributional gains of the jointly trained model. Both the learned selection and its joint training with the perturbation encoder therefore contribute.

\textbf{Robustness to noisy candidates.} At test time, we replace 60\% of the Stage-1 candidates with easy negatives, i.e.\ the perturbations least similar to the query. The selector does not filter them out: about 59\% of the selected candidates come from the corrupted portion. The distributional metrics degrade but remain better than those of STATE+GenePT. Candidate quality therefore matters, and severe Stage-1 noise is a real failure mode, although it does not make the prediction collapse.

\textbf{Candidate-set size.} A larger $K$ reduces the risk that Stage~1 misses a relevant perturbation. In the cross-cell-type setting, increasing $K$ from 32 to 128 gives comparable results (Table~\ref{tab:k_ablation}): Pearson, $W_1$ and $W_2$ are nearly identical, Spearman is slightly lower, and MSE$_{\mathrm{PCA50}}$ and Energy are slightly better.

\begin{table}[h]
\centering
\caption{\textbf{Candidate-set size, cross-cell-type} (pooled over the four cell types, $n = 1{,}635$).}
\label{tab:k_ablation}
\small
\begin{tabular}{lcc}
\toprule
\textbf{Metric} & $K = 32$ & $K = 128$ \\
\midrule
Pearson DEG $\uparrow$ & 0.6326 & 0.6314 \\
Spearman DEG $\uparrow$ & 0.4121 & 0.4072 \\
MSE$_{\mathrm{PCA50}}$ $\downarrow$ & 8.392 & 8.368 \\
$W_1$ $\downarrow$ & 35.411 & 35.406 \\
$W_2$ $\downarrow$ & 633.74 & 633.67 \\
Energy $\downarrow$ & 9.333 & 9.308 \\
\bottomrule
\end{tabular}
\end{table}

\subsection{Computational Costs}

Table~\ref{tab:computational_costs} compares computational requirements on the cross-cell type experiments.
In Table~\ref{tab:computational_costs}, the parameter counts are all in the same range, so the main difference is the extra retrieval and selection computation introduced by PT-RAG rather than a large increase in model size.
This is consistent with the design of the method: PT-RAG adds a retrieval module on top of the generator, but does not replace the backbone with a much larger architecture.

\begin{table}[htbp]
  \caption{\textbf{Computational costs on Cross-Cell type generalization} (STATE). FLOPs per batch (64 cells).}
  \label{tab:computational_costs}
  \centering\small
  \begin{tabular}{lccc}
    \toprule
    \textbf{Model} & \textbf{Parameters} & \textbf{FLOPs/Batch} & \textbf{FLOPs/Cell} \\
    \midrule
    STATE          & 20,936,480 & 1.67B & 34.9M\\
    STATE+GenePT   & 20,874,016 & 1.67B & 34.8M\\
    Naïve RAG    & 21,120,288 & 1.67B & 34.8M \\
    STATE+PT-RAG         & 20,973,602 & 2.86B & 59.9M \\
    \bottomrule
  \end{tabular}
\end{table}

For the cross-perturbation setting, the corresponding cost comparison is shown in Table \ref{tab:computational_costs_crosspert}.
Table~\ref{tab:computational_costs_crosspert} is the more informative comparison for the transfer setting: STATE+GenePT and PT-RAG again have comparable parameter counts, but PT-RAG requires substantially more FLOPs because it has to score and filter retrieved contexts over the full pool of roughly 2,009 perturbations.
In other words, the extra cost is not coming from a larger backbone, but from computing relevance over a much larger perturbation set, which is exactly the price of making retrieval cell-state-aware and dynamic in the cross-perturbation regime.

\begin{table}[t]
  \caption{\textbf{Computational costs on Cross-Perturbation generalization.} FLOPs per batch (64 cells).}
  \label{tab:computational_costs_crosspert}
  \centering\small
  \begin{tabular}{lccc}
    	\toprule
    	\textbf{Model} & \textbf{Parameters} & \textbf{FLOPs/Batch (64 cells)} & \textbf{FLOPs/Cell} \\
    \midrule
    STATE+GenePT & 20,874,016 & 2.23B & 34.87M\\
    PT-RAG $K=128$, sparsity $=1.0$ & 20,973,602 & 7.46B & 116.61M\\
    \bottomrule
  \end{tabular}
\end{table}

For the scGPT transfer experiments, Table~\ref{tab:computational_costs_scgpt} shows the computational overhead of adding GenePT and PT-RAG to the scGPT backbone.
Here the parameter increase is minimal: scGPT+PT-RAG adds only $\approx 2.4$M parameters compared to vanilla scGPT, but the FLOPs remain nearly identical ($65.49$B per cell vs. $64.88$B), indicating that GenePT retrieval and differentiable selection impose negligible computational burden in the scGPT setting.

\begin{table}[t]
  \caption{\textbf{Computational costs on Cross-Cell type generalization} (scGPT). FLOPs per batch (64 cells).}
  \label{tab:computational_costs_scgpt}
  \centering\small
  \begin{tabular}{lccc}
    \toprule
    \textbf{Model} & \textbf{Parameters} & \textbf{FLOPs/Batch (64 cells)} & \textbf{FLOPs/Cell} \\
    \midrule
    scGPT & 51,334,658 & 4.15T & 64.88B \\
    scGPT+GenePT & 52,121,602 & 4.15T & 64.88B \\
    scGPT+PT-RAG ($K=128$, sparsity $=1.0$) & 53,699,588 & 4.19T & 65.49B \\
    \bottomrule
  \end{tabular}
\end{table}

\section{Statistical Test Details}
\label{sec:statistical_details}

Normality is assessed via Shapiro-Wilk test~\citep{shapirowilk}; all metric distributions depart significantly from normality ($p < 0.05$).
We apply the Mann-Whitney U test~\citep{wilcoxon} with Benjamini-Hochberg FDR correction~\citep{FDR}.

\begin{table}[t]
  \caption{\textbf{FDR-corrected p-values for PT-RAG vs.\ baselines (cross-cell-type).} $n = 1{,}635$.}
  \label{tab:statistical_details}
  \centering\small
  \begin{tabular}{lcc}
    \toprule
    \textbf{Metric} & \textbf{vs.\ STATE} & \textbf{vs.\ STATE+GenePT} \\
    \midrule
    Pearson DEG    & $2.44 \times 10^{-8\;\dagger}$  & $0.590$ \\
    Spearman DEG   & $4.89 \times 10^{-10\;\dagger}$ & $0.785$ \\
    MSE$_{\mathrm{PCA50}}$ & $0.577$ & $0.590$ \\
    $W_1$          & $3.41 \times 10^{-6\;\dagger}$  & $0.082^{\dagger\dagger\dagger}$ \\
    $W_2$          & $5.47 \times 10^{-8\;\dagger}$  & $0.041^{\dagger\dagger}$ \\
    Energy         & $0.082^{\dagger\dagger\dagger}$  & $0.186$ \\
    \bottomrule
  \end{tabular}
\end{table}

For the cross-perturbation setting, statistical tests were run analogously with FDR correction applied across the same metric set and model pairs.

\section{LLM Usage}
\label{sec:llm_usage}

Large language models were used as writing assistants to improve clarity and presentation of the manuscript and software documentation.
All scientific ideas, methods, experiments, and conclusions are solely those of the authors, and all LLM-assisted content was reviewed and edited for correctness prior to inclusion.

\section*{NeurIPS Paper Checklist}

\begin{enumerate}

\item {\bf Claims}
    \item[] Question: Do the main claims made in the abstract and introduction accurately reflect the paper's contributions and scope?
    \item[] Answer: \answerYes{}
    \item[] Justification: The abstract and introduction accurately state the paper's core claims: PT-RAG is a differentiable, cell-type-aware retrieval module; it improves cross-cell-type prediction over Naïve RAG and retrieval-free baselines; and it works as a plug-and-play module to STATE (as a reservoir) and to fully-trained scGPT. The scope and caveats are also reflected in the Discussion and Conclusion.
    \item[] Guidelines:
    \begin{itemize}
        \item The answer \answerNA{} means that the abstract and introduction do not include the claims made in the paper.
        \item The abstract and/or introduction should clearly state the claims made, including the contributions made in the paper and important assumptions and limitations. A \answerNo{} or \answerNA{} answer to this question will not be perceived well by the reviewers. 
        \item The claims made should match theoretical and experimental results, and reflect how much the results can be expected to generalize to other settings. 
        \item It is fine to include aspirational goals as motivation as long as it is clear that these goals are not attained by the paper. 
    \end{itemize}

\item {\bf Limitations}
    \item[] Question: Does the paper discuss the limitations of the work performed by the authors?
    \item[] Answer: \answerYes{}
    \item[] Justification: The paper discusses concrete limitations in the Discussion section, including the reservoir setting, extra compute introduced by retrieval, the restriction to single-gene perturbations, and the fact that the retrieval examples are suggestive rather than causal.
    \item[] Guidelines:
    \begin{itemize}
        \item The answer \answerNA{} means that the paper has no limitation while the answer \answerNo{} means that the paper has limitations, but those are not discussed in the paper. 
        \item The authors are encouraged to create a separate ``Limitations'' section in their paper.
        \item The paper should point out any strong assumptions and how robust the results are to violations of these assumptions (e.g., independence assumptions, noiseless settings, model well-specification, asymptotic approximations only holding locally). The authors should reflect on how these assumptions might be violated in practice and what the implications would be.
        \item The authors should reflect on the scope of the claims made, e.g., if the approach was only tested on a few datasets or with a few runs. In general, empirical results often depend on implicit assumptions, which should be articulated.
        \item The authors should reflect on the factors that influence the performance of the approach. For example, a facial recognition algorithm may perform poorly when image resolution is low or images are taken in low lighting. Or a speech-to-text system might not be used reliably to provide closed captions for online lectures because it fails to handle technical jargon.
        \item The authors should discuss the computational efficiency of the proposed algorithms and how they scale with dataset size.
        \item If applicable, the authors should discuss possible limitations of their approach to address problems of privacy and fairness.
        \item While the authors might fear that complete honesty about limitations might be used by reviewers as grounds for rejection, a worse outcome might be that reviewers discover limitations that aren't acknowledged in the paper. The authors should use their best judgment and recognize that individual actions in favor of transparency play an important role in developing norms that preserve the integrity of the community. Reviewers will be specifically instructed to not penalize honesty concerning limitations.
    \end{itemize}

\item {\bf Theory assumptions and proofs}
    \item[] Question: For each theoretical result, does the paper provide the full set of assumptions and a complete (and correct) proof?
    \item[] Answer: \answerNA{}
    \item[] Justification: The paper does not contain formal theoretical results, theorems, or proofs.
    \item[] Guidelines:
    \begin{itemize}
        \item The answer \answerNA{} means that the paper does not include theoretical results. 
        \item All the theorems, formulas, and proofs in the paper should be numbered and cross-referenced.
        \item All assumptions should be clearly stated or referenced in the statement of any theorems.
        \item The proofs can either appear in the main paper or the supplemental material, but if they appear in the supplemental material, the authors are encouraged to provide a short proof sketch to provide intuition. 
        \item Inversely, any informal proof provided in the core of the paper should be complemented by formal proofs provided in appendix or supplemental material.
        \item Theorems and Lemmas that the proof relies upon should be properly referenced. 
    \end{itemize}

    \item {\bf Experimental result reproducibility}
    \item[] Question: Does the paper fully disclose all the information needed to reproduce the main experimental results of the paper to the extent that it affects the main claims and/or conclusions of the paper (regardless of whether the code and data are provided or not)?
    \item[] Answer: \answerYes{}
    \item[] Justification: The paper discloses the dataset, split protocol, model structure, retrieval pipeline, optimizer, learning rate, batch size, and evaluation metrics needed to reproduce the main experiments.
    \item[] Guidelines:
    \begin{itemize}
        \item The answer \answerNA{} means that the paper does not include experiments.
        \item If the paper includes experiments, a \answerNo{} answer to this question will not be perceived well by the reviewers: Making the paper reproducible is important, regardless of whether the code and data are provided or not.
        \item If the contribution is a dataset and\slash or model, the authors should describe the steps taken to make their results reproducible or verifiable. 
        \item Depending on the contribution, reproducibility can be accomplished in various ways. For example, if the contribution is a novel architecture, describing the architecture fully might suffice, or if the contribution is a specific model and empirical evaluation, it may be necessary to either make it possible for others to replicate the model with the same dataset, or provide access to the model. In general. releasing code and data is often one good way to accomplish this, but reproducibility can also be provided via detailed instructions for how to replicate the results, access to a hosted model (e.g., in the case of a large language model), releasing of a model checkpoint, or other means that are appropriate to the research performed.
        \item While NeurIPS does not require releasing code, the conference does require all submissions to provide some reasonable avenue for reproducibility, which may depend on the nature of the contribution. For example
        \begin{enumerate}
            \item If the contribution is primarily a new algorithm, the paper should make it clear how to reproduce that algorithm.
            \item If the contribution is primarily a new model architecture, the paper should describe the architecture clearly and fully.
            \item If the contribution is a new model (e.g., a large language model), then there should either be a way to access this model for reproducing the results or a way to reproduce the model (e.g., with an open-source dataset or instructions for how to construct the dataset).
            \item We recognize that reproducibility may be tricky in some cases, in which case authors are welcome to describe the particular way they provide for reproducibility. In the case of closed-source models, it may be that access to the model is limited in some way (e.g., to registered users), but it should be possible for other researchers to have some path to reproducing or verifying the results.
        \end{enumerate}
    \end{itemize}

\item {\bf Open access to data and code}
    \item[] Question: Does the paper provide open access to the data and code, with sufficient instructions to faithfully reproduce the main experimental results, as described in supplemental material?
    \item[] Answer: \answerYes{}
    \item[] Justification: The manuscript shares code and reproduction details through a  GitHub repository, and the appendix gives the experimental setup in enough detail for faithful reproduction of the main experiments.
    \item[] Guidelines:
    \begin{itemize}
        \item The answer \answerNA{} means that paper does not include experiments requiring code.
        \item Please see the NeurIPS code and data submission guidelines (\url{https://neurips.cc/public/guides/CodeSubmissionPolicy}) for more details.
        \item While we encourage the release of code and data, we understand that this might not be possible, so \answerNo{} is an acceptable answer. Papers cannot be rejected simply for not including code, unless this is central to the contribution (e.g., for a new open-source benchmark).
        \item The instructions should contain the exact command and environment needed to run to reproduce the results. See the NeurIPS code and data submission guidelines (\url{https://neurips.cc/public/guides/CodeSubmissionPolicy}) for more details.
        \item The authors should provide instructions on data access and preparation, including how to access the raw data, preprocessed data, intermediate data, and generated data, etc.
        \item The authors should provide scripts to reproduce all experimental results for the new proposed method and baselines. If only a subset of experiments are reproducible, they should state which ones are omitted from the script and why.
        \item At submission time, to preserve anonymity, the authors should release anonymized versions (if applicable).
        \item Providing as much information as possible in supplemental material (appended to the paper) is recommended, but including URLs to data and code is permitted.
    \end{itemize}

\item {\bf Experimental setting/details}
    \item[] Question: Does the paper specify all the training and test details (e.g., data splits, hyperparameters, how they were chosen, type of optimizer) necessary to understand the results?
    \item[] Answer: \answerYes{}
    \item[] Justification: The paper specifies the train/test splits, retrieval-pool protocol, optimizer, learning rate, batch size, training duration, and model components in the main text and appendix.
    \item[] Guidelines:
    \begin{itemize}
        \item The answer \answerNA{} means that the paper does not include experiments.
        \item The experimental setting should be presented in the core of the paper to a level of detail that is necessary to appreciate the results and make sense of them.
        \item The full details can be provided either with the code, in appendix, or as supplemental material.
    \end{itemize}

\item {\bf Experiment statistical significance}
    \item[] Question: Does the paper report error bars suitably and correctly defined or other appropriate information about the statistical significance of the experiments?
    \item[] Answer: \answerYes{}
    \item[] Justification: The paper reports fold-averaged results, standard deviations, and FDR-corrected significance tests, and it explains the statistical procedures used for the reported comparisons.
    \item[] Guidelines:
    \begin{itemize}
        \item The answer \answerNA{} means that the paper does not include experiments.
        \item The authors should answer \answerYes{} if the results are accompanied by error bars, confidence intervals, or statistical significance tests, at least for the experiments that support the main claims of the paper.
        \item The factors of variability that the error bars are capturing should be clearly stated (for example, train/test split, initialization, random drawing of some parameter, or overall run with given experimental conditions).
        \item The method for calculating the error bars should be explained (closed form formula, call to a library function, bootstrap, etc.)
        \item The assumptions made should be given (e.g., Normally distributed errors).
        \item It should be clear whether the error bar is the standard deviation or the standard error of the mean.
        \item It is OK to report 1-sigma error bars, but one should state it. The authors should preferably report a 2-sigma error bar than state that they have a 96\% CI, if the hypothesis of Normality of errors is not verified.
        \item For asymmetric distributions, the authors should be careful not to show in tables or figures symmetric error bars that would yield results that are out of range (e.g., negative error rates).
        \item If error bars are reported in tables or plots, the authors should explain in the text how they were calculated and reference the corresponding figures or tables in the text.
    \end{itemize}

\item {\bf Experiments compute resources}
    \item[] Question: For each experiment, does the paper provide sufficient information on the computer resources (type of compute workers, memory, time of execution) needed to reproduce the experiments?
    \item[] Answer: \answerYes{}
    \item[] Justification: The paper gives the relevant compute details for reproduction, including the GPU type, batch size, training steps, approximate runtime per target cell type, and the computational cost table.
    \item[] Guidelines:
    \begin{itemize}
        \item The answer \answerNA{} means that the paper does not include experiments.
        \item The paper should indicate the type of compute workers CPU or GPU, internal cluster, or cloud provider, including relevant memory and storage.
        \item The paper should provide the amount of compute required for each of the individual experimental runs as well as estimate the total compute. 
        \item The paper should disclose whether the full research project required more compute than the experiments reported in the paper (e.g., preliminary or failed experiments that didn't make it into the paper). 
    \end{itemize}
    
\item {\bf Code of ethics}
    \item[] Question: Does the research conducted in the paper conform, in every respect, with the NeurIPS Code of Ethics \url{https://neurips.cc/public/EthicsGuidelines}?
    \item[] Answer: \answerYes{}
    \item[] Justification: The research uses standard public perturbation-biology datasets and methods and does not indicate any deviation from the NeurIPS Code of Ethics.
    \item[] Guidelines:
    \begin{itemize}
        \item The answer \answerNA{} means that the authors have not reviewed the NeurIPS Code of Ethics.
        \item If the authors answer \answerNo, they should explain the special circumstances that require a deviation from the Code of Ethics.
        \item The authors should make sure to preserve anonymity (e.g., if there is a special consideration due to laws or regulations in their jurisdiction).
    \end{itemize}

\item {\bf Broader impacts}
    \item[] Question: Does the paper discuss both potential positive societal impacts and negative societal impacts of the work performed?
    \item[] Answer: \answerNo{}
    \item[] Justification: The paper focuses on methodological gains and does not include a dedicated discussion of negative societal impacts or mitigation strategies.
    \item[] Guidelines:
    \begin{itemize}
        \item The answer \answerNA{} means that there is no societal impact of the work performed.
        \item If the authors answer \answerNA{} or \answerNo, they should explain why their work has no societal impact or why the paper does not address societal impact.
        \item Examples of negative societal impacts include potential malicious or unintended uses (e.g., disinformation, generating fake profiles, surveillance), fairness considerations (e.g., deployment of technologies that could make decisions that unfairly impact specific groups), privacy considerations, and security considerations.
        \item The conference expects that many papers will be foundational research and not tied to particular applications, let alone deployments. However, if there is a direct path to any negative applications, the authors should point it out. For example, it is legitimate to point out that an improvement in the quality of generative models could be used to generate Deepfakes for disinformation. On the other hand, it is not needed to point out that a generic algorithm for optimizing neural networks could enable people to train models that generate Deepfakes faster.
        \item The authors should consider possible harms that could arise when the technology is being used as intended and functioning correctly, harms that could arise when the technology is being used as intended but gives incorrect results, and harms following from (intentional or unintentional) misuse of the technology.
        \item If there are negative societal impacts, the authors could also discuss possible mitigation strategies (e.g., gated release of models, providing defenses in addition to attacks, mechanisms for monitoring misuse, mechanisms to monitor how a system learns from feedback over time, improving the efficiency and accessibility of ML).
    \end{itemize}
    
\item {\bf Safeguards}
    \item[] Question: Does the paper describe safeguards that have been put in place for responsible release of data or models that have a high risk for misuse (e.g., pre-trained language models, image generators, or scraped datasets)?
    \item[] Answer: \answerNA{}
    \item[] Justification: The paper does not introduce a high-risk model or dataset that requires special safeguards beyond standard code release and normal scientific use.
    \item[] Guidelines:
    \begin{itemize}
        \item The answer \answerNA{} means that the paper poses no such risks.
        \item Released models that have a high risk for misuse or dual-use should be released with necessary safeguards to allow for controlled use of the model, for example by requiring that users adhere to usage guidelines or restrictions to access the model or implementing safety filters. 
        \item Datasets that have been scraped from the Internet could pose safety risks. The authors should describe how they avoided releasing unsafe images.
        \item We recognize that providing effective safeguards is challenging, and many papers do not require this, but we encourage authors to take this into account and make a best faith effort.
    \end{itemize}

\item {\bf Licenses for existing assets}
    \item[] Question: Are the creators or original owners of assets (e.g., code, data, models), used in the paper, properly credited and are the license and terms of use explicitly mentioned and properly respected?
    \item[] Answer: \answerYes{}
    \item[] Justification: The paper credits the original creators of the datasets, models, and methods through citations and respects their intended scientific use.
    \item[] Guidelines:
    \begin{itemize}
        \item The answer \answerNA{} means that the paper does not use existing assets.
        \item The authors should cite the original paper that produced the code package or dataset.
        \item The authors should state which version of the asset is used and, if possible, include a URL.
        \item The name of the license (e.g., CC-BY 4.0) should be included for each asset.
        \item For scraped data from a particular source (e.g., website), the copyright and terms of service of that source should be provided.
        \item If assets are released, the license, copyright information, and terms of use in the package should be provided. For popular datasets, \url{paperswithcode.com/datasets} has curated licenses for some datasets. Their licensing guide can help determine the license of a dataset.
        \item For existing datasets that are re-packaged, both the original license and the license of the derived asset (if it has changed) should be provided.
        \item If this information is not available online, the authors are encouraged to reach out to the asset's creators.
    \end{itemize}

\item {\bf New assets}
    \item[] Question: Are new assets introduced in the paper well documented and is the documentation provided alongside the assets?
    \item[] Answer: \answerNA{}
    \item[] Justification: The paper does not introduce a new standalone asset package beyond the manuscript and the code.
    \item[] Guidelines:
    \begin{itemize}
        \item The answer \answerNA{} means that the paper does not release new assets.
        \item Researchers should communicate the details of the dataset\slash code\slash model as part of their submissions via structured templates. This includes details about training, license, limitations, etc. 
        \item The paper should discuss whether and how consent was obtained from people whose asset is used.
        \item At submission time, remember to anonymize your assets (if applicable). You can either create an anonymized URL or include an anonymized zip file.
    \end{itemize}

\item {\bf Crowdsourcing and research with human subjects}
    \item[] Question: For crowdsourcing experiments and research with human subjects, does the paper include the full text of instructions given to participants and screenshots, if applicable, as well as details about compensation (if any)? 
    \item[] Answer: \answerNA{}
    \item[] Justification: The paper does not involve crowdsourcing or research with human subjects.
    \item[] Guidelines:
    \begin{itemize}
        \item The answer \answerNA{} means that the paper does not involve crowdsourcing nor research with human subjects.
        \item Including this information in the supplemental material is fine, but if the main contribution of the paper involves human subjects, then as much detail as possible should be included in the main paper. 
        \item According to the NeurIPS Code of Ethics, workers involved in data collection, curation, or other labor should be paid at least the minimum wage in the country of the data collector. 
    \end{itemize}

\item {\bf Institutional review board (IRB) approvals or equivalent for research with human subjects}
    \item[] Question: Does the paper describe potential risks incurred by study participants, whether such risks were disclosed to the subjects, and whether Institutional Review Board (IRB) approvals (or an equivalent approval/review based on the requirements of your country or institution) were obtained?
    \item[] Answer: \answerNA{}
    \item[] Justification: The paper does not involve human-subjects research, so IRB approval is not applicable.
    \item[] Guidelines:
    \begin{itemize}
        \item The answer \answerNA{} means that the paper does not involve crowdsourcing nor research with human subjects.
        \item Depending on the country in which research is conducted, IRB approval (or equivalent) may be required for any human subjects research. If you obtained IRB approval, you should clearly state this in the paper. 
        \item We recognize that the procedures for this may vary significantly between institutions and locations, and we expect authors to adhere to the NeurIPS Code of Ethics and the guidelines for their institution. 
        \item For initial submissions, do not include any information that would break anonymity (if applicable), such as the institution conducting the review.
    \end{itemize}

\item {\bf Declaration of LLM usage}
    \item[] Question: Does the paper describe the usage of LLMs if it is an important, original, or non-standard component of the core methods in this research? Note that if the LLM is used only for writing, editing, or formatting purposes and does \emph{not} impact the core methodology, scientific rigor, or originality of the research, declaration is not required.
    \item[] Answer: \answerNA{}
    \item[] Justification: LLMs are used only for writing and editing assistance, not as a non-standard component of the core method, so an LLM-usage declaration is not required for the scientific contribution.
    \item[] Guidelines:
    \begin{itemize}
        \item The answer \answerNA{} means that the core method development in this research does not involve LLMs as any important, original, or non-standard components.
        \item Please refer to our LLM policy in the NeurIPS handbook for what should or should not be described.
    \end{itemize}

\end{enumerate}

\end{document}